\documentclass[journal]{IEEEtran}

\usepackage{amsmath}
\usepackage{amsthm}
\usepackage{amssymb}
\usepackage{amsbsy}
\usepackage{graphicx}
\usepackage{subfigure}
\usepackage{multirow}
\usepackage{epsfig}
\usepackage{float}
\usepackage{cite}
\usepackage{indentfirst}
\usepackage{epsfig}
\usepackage{color}
\usepackage{multirow}
\usepackage{url}  
\usepackage{color}
\usepackage{times}
\usepackage{soul}
\usepackage{url}
\usepackage{algorithm}
\usepackage{algorithmic}
\usepackage{times}
\usepackage{epsfig}
\usepackage{amsmath}
\usepackage{amssymb}
\usepackage{multirow}
\usepackage{diagbox}

\newtheorem*{theorem*}{Theorem}

\newtheorem*{corollary*}{Corollary}

\newtheorem*{lemma*}{Lemma}

\newtheorem*{proposition*}{Proposition}

\makeatletter

\newcommand{\Rmnum}[1]{\expandafter\@slowromancap\romannumeral #1@}
\makeatother

\usepackage{color}

\ifCLASSINFOpdf

\else

\fi

\begin{document}

\title{Self-supervised Training of Graph Convolutional Networks}

\author{Qikui Zhu, Bo Du,~\IEEEmembership{Senior Member,~IEEE}, Pingkun~Yan,~\IEEEmembership{Senior Member,~IEEE}
%
\thanks{
Q.~Zhu is with School of Computer Science,
Wuhan University, Wuhan, China. (e-mail: QikuiZhu@whu.edu.cn)} 
%
\thanks{B.~Du is with School of Computer Science, Wuhan University, Wuhan, China. (e-mail: dubo@whu.edu.cn)}%
\thanks{P.~Yan is with the Department of Biomedical Engineering and the Center for Biotechnology and Interdisciplinary Studies at Rensselaer Polytechnic Institute, Troy, NY, USA 12180. (e-mail: yanp2@rpi.edu)}%
}

\maketitle

\begin{abstract}
Graph Convolutional Networks (GCNs) have been successfully applied to analyze non-grid data, where the classical convolutional neural networks (CNNs) cannot be directly used.
One similarity shared by GCNs and CNNs is the requirement of massive amount of labeled data for network training.
In addition, GCNs need the adjacency matrix as input to define the relationship between those non-grid data, which leads to all of data including training, validation and test data typically forms only one graph structures data for training.
Furthermore, the adjacency matrix is usually pre-defined and stationary, which makes the data augmentation strategies cannot be employed on the constructed graph structures data to augment the amount of training data. To further improve the learning capacity and model performance under the limited training data, in this paper, we propose two types of self-supervised learning strategies to exploit available information from the input graph structure data itself. Our proposed self-supervised learning strategies are examined on two representative GCN models with three public citation network datasets - Citeseer, Cora and Pubmed. The experimental results demonstrate the generalization ability as well as the portability of our proposed strategies, which can significantly improve the performance of GCNs with the power of self-supervised learning in improving feature learning.
\end{abstract}

\begin{IEEEkeywords}
Graph Convolutional Networks, self-supervised learning, feature learning.
\end{IEEEkeywords}

\IEEEpeerreviewmaketitle

\section{Introduction}

\IEEEPARstart{T}{he} recent deep learning renaissance started with Convolutional Neural Networks (CNNs), which have achieved revolutionary performance in many fields, including computer vision, natural language processing, and medical image computing, owning to their capacity of learning discriminating features.
Driven by the success of CNNs in the computer vision domain, there has been increased interest in applying deep learning, especially Graph Convolutional Networks (GCNs), to arbitrarily structured data such as social networks, knowledge graphs and chemical molecules.
For example, Marino et al.~\cite{marino2017more} presented a Graph Search Neural Network (GSNN), which can efficiently incorporate potentially large knowledge graphs for use as extra information to improve image classification. Wang et al.~\cite{wang2018zero} introduced GCNs into the problem of zero-shot recognition. They proposed using the semantic embeddings of a category and the knowledge graph that encodes the relationship of the novel category to familiar categories. Moreover, GCNs can also be used in medical image field. For example, Parisot et al.~\cite{parisot2017spectral} proposed a novel concept of graph convolutions for population-based brain analysis. To enable utilization of both image and non-image information, they first constructed a population graph by combining image-based patient-specific information with non-imaging-based pairwise interactions, then used this structure to train a GCN for the semi-supervised classification of populations. Although significant progress has been achieved in above areas, a huge room are still waiting to be promoted.

However, the major factor limiting convolutional neural networks based models performance is the amount of annotated data.
GCNs are not exception, which also require massive amounts of manually labeled data during training.
What's more, GCNs meet greater challenges compared with CNNs. Taking two representative GCN models - GCN~\cite{kipf2016semi}, GAT~\cite{velivckovic2017graph} - as examples. In the experiment, all of data including training, validation and testing data are used to construct the input graph structures data. Moreover, the data augmentation strategies, such as rotation, flipping and cropping, cannot be employed on graph structures data, which leads to training data has only one sample. Even though both GCN and GAT have only two layers and fewer filters, the overfitting problem still cannot be overcome. Furthermore, the capacity of learning discriminating of those models also be limited, owning to only one training sample can be used.
Therefore, making the graph convolutional neural networks based models learning without requiring manual annotated data is of crucial importance in further improving model performance , particularly for GCNs.

To tackle the above mentioned challenges in GCNs, in this paper, we propose to exploit available information from the input graph structures data itself to improve the capacity of learning discriminating and the performance of graph based models. Different from regular grid data, such as images or videos, the graph structures use edges to store the structure information and also the relationship between nodes. Aside from the structure information, each node inside graph also contains a feature vector that stores the node representation. To utilize those information, two types of self-supervised learning strategies: Randomly Removing Links (RRL) and Randomly Covering Features (RCF) are proposed in this paper. Our proposed self-supervised learning (SSL) strategies does not require any labeled information for feature learning, and instead uses only the information provided by the graph structures data itself. Those information can provide a surrogate supervision signal for feature learning.

To the best of our knowledge, this is the first work employing SSL for training GCNs. To evaluate the effectiveness, generalization as well as portability of SSL in improving the performance of GCNs, two representative GCN models, GCN~\cite{kipf2016semi} and GAT~\cite{velivckovic2017graph} with  three public citation network datasets - Citeseer, Cora and Pubmed~\cite{yang2016revisiting} - are used to evaluate our proposed SSL strategies. The results corroborate the effectiveness of our proposed SSL strategies, which achieve clearly improved performance compared with other state-of-the-art methods.


\begin{figure*}
  \centering
  \includegraphics[width=0.9\textwidth]{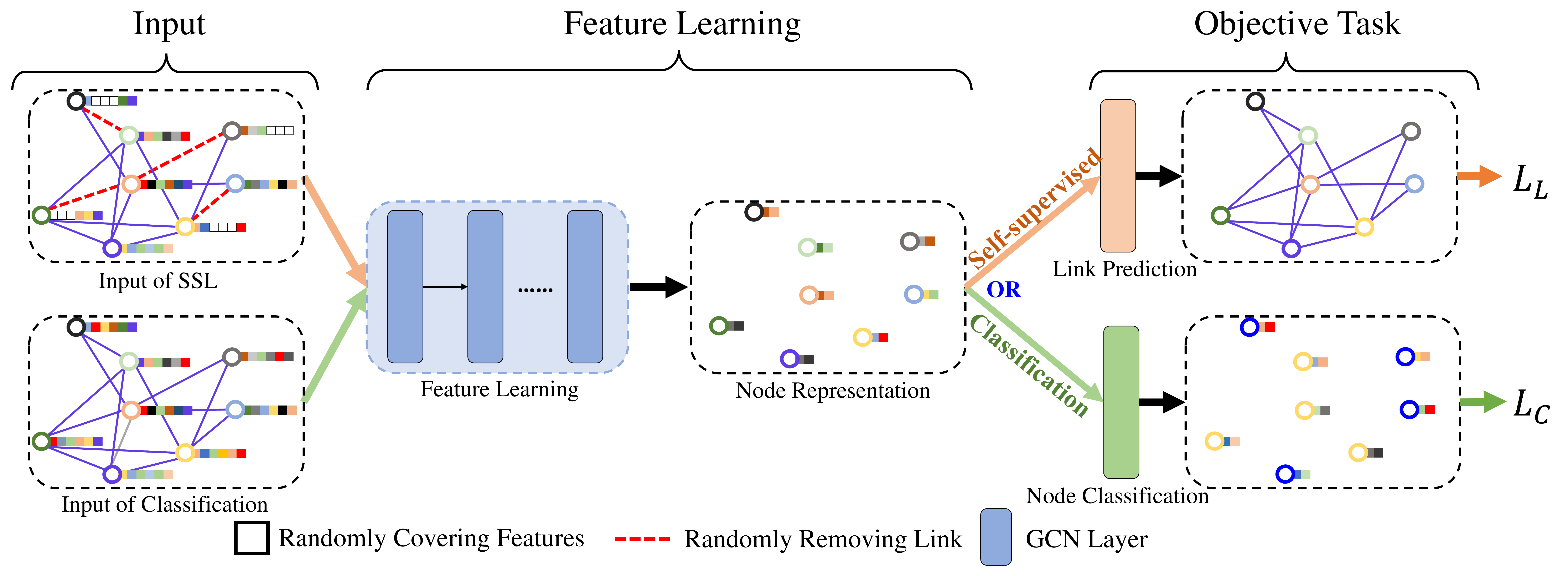}
  \caption{Overview of the proposed self-supervised training of graph convolutional networks.}\label{ProposedModel}
\end{figure*}

\section{Related Work}

\subsection{Graph Convolutional Networks}
%
Graph Convolutional Neural Networks (GCNs) have recently become one of the most powerful tools for graph analytics tasks in many applications, including molecule generation, recommendation systems, anomaly detection and citation network. Based on the definition domain of GCNs, GCNs-based methods can be categorized into spectral convolution and spatial convolution methods. In the former, Bruna~et al.~\cite{bruna2013spectral} first developed the spectral convolution neural network (Spectral CNN) based on spectral graph theory. However, the operation of Spectral CNN requires multiple with the eigenvector matrix and computing the eigendecomposition of the graph Laplacian matrix, which may be prohibitively expensive for large graphs and make the computation of Spectral CNNs expensive. To circumvent this challenge, Defferrard et~al.~\cite{defferrard2016convolutional} proposed ChebNet, which defines a filter as Chebyshev polynomials of the diagonal matrix of eigenvalues. Inside the ChebNet, moreover, the spectral graph convolution operator can be well approximated by a truncated expansion in terms of Chebyshev polynomials. To further alleviate overfitting problem on local neighborhood structures for graphs with very wide node degree distributions, Kipf et~al.~\cite{kipf2016semi} proposed a first-order approximation of ChebNet. These authors limited the layer-wise convolution operation and made the ChebNet becomes a linear function on the graph Laplacian spectrum. Therefore, constraining the number of parameters, avoiding numerical instabilities and exploding or vanishing gradients bring by stacking multiple graph convolutional layers.

Unlike methods based on spectral GCNs, spatial GCNs-based methods define graph convolution based on the node’s spatial relations. Taking image data as an example, the whole image can be considered as a special type of graph, where each pixel represents a node and adjacent nodes denote neighborhoods. Same like conventional CNNs, the spatial graph convolution makes each node learn and update node representation from neighborhoods. For example, Duvenaud et~al.~\cite{duvenaud2015convolutional} proposed an end-to-end learning convolutional neural network, which can operate directly on graph data with arbitrary size and shape. Atwood et~al.\cite{atwood2016diffusion} extended the convolutional neural networks to diffusion-convolutional neural networks (DCNNs) for general graph-structured data by introducing a `diffusion-convolution' operation. The proposed DCNNs can learn a representation that encapsulates the results of graph diffusion and further incorporates the contextual information of each node in the graph.
Monti et~al.~\cite{monti2017geometric} presented a unified mixture model network (MoNet) that formulates convolution-like operations as template matching with local intrinsic patches on graphs or manifolds, which allows to generalize CNNs architectures to graph structed data as well as for the learning of, stationary and compositional task-specific features.
Another representative spatial GCN is Graph Attention Networks (GATs), a novel convolution-style neural network that operates on graph-structured data and leverages masked self-attentional layers, which was first proposed by ~\cite{velivckovic2017graph}.
\subsection{Self-supervised Learning}
In recent years, there has been significant interest in learning meaningful representations using self-supervised learning (SSL), with the aim of overcoming the limitation that deep learning based models require requires large amounts of labelled data. Unlike supervised or semi-supervised learning, SSL does not require any labeled data, as it uses only the visual information provided by the images or video itself; this information can provide a surrogate supervision signal for feature learning. Based on the type of pretext task used in the SSL, we can divide the types of self-supervised models into different categories: image rotation, relative position, colorization and image inpainting.
For example, Doersch et al.~\cite{doersch2015unsupervised} proposed an unsupervised visual representation learning method based on context prediction, which extracts random pairs of patches from each image and trains a CNN to predict the position of the second patch relative to the first. Experimental results have demonstrated that the feature representation learned using this within-image context is capable of capturing visual similarity across images.
Gidaris et al.~\cite{gidaris2018unsupervised} introduced the geometric transformations into SSL and proven that image features could be learned by training ConvNets to recognize the 2D rotation applied to the image, which significantly narrows the gap between unsupervised and supervised feature learning. Zhang et al.~ \cite{zhang2016colorful} treated colorization as a form of self-supervised representation learning. Given a grayscale photograph, these authors tried to produce vibrant and realistic colorizations. Moreover, driven by context-based pixel prediction, Pathak et al.~\cite{pathak2016context} presented an unsupervised visual feature learning model, which predicates the missing parts of a scene from their surroundings. During training process, this model can learn a representation that captures not only the appearance but also the semantics of visual structures.
Chen et al.~\cite{CHEN2019101539} proposed a novel SSL strategy, known as context restoration, allowing it to better exploit unlabeled images. The task of context restoration involves predicting the relative positions of image patches and local context prediction.

\begin{figure}[t]
  \centering
  \includegraphics[width=0.8\columnwidth]{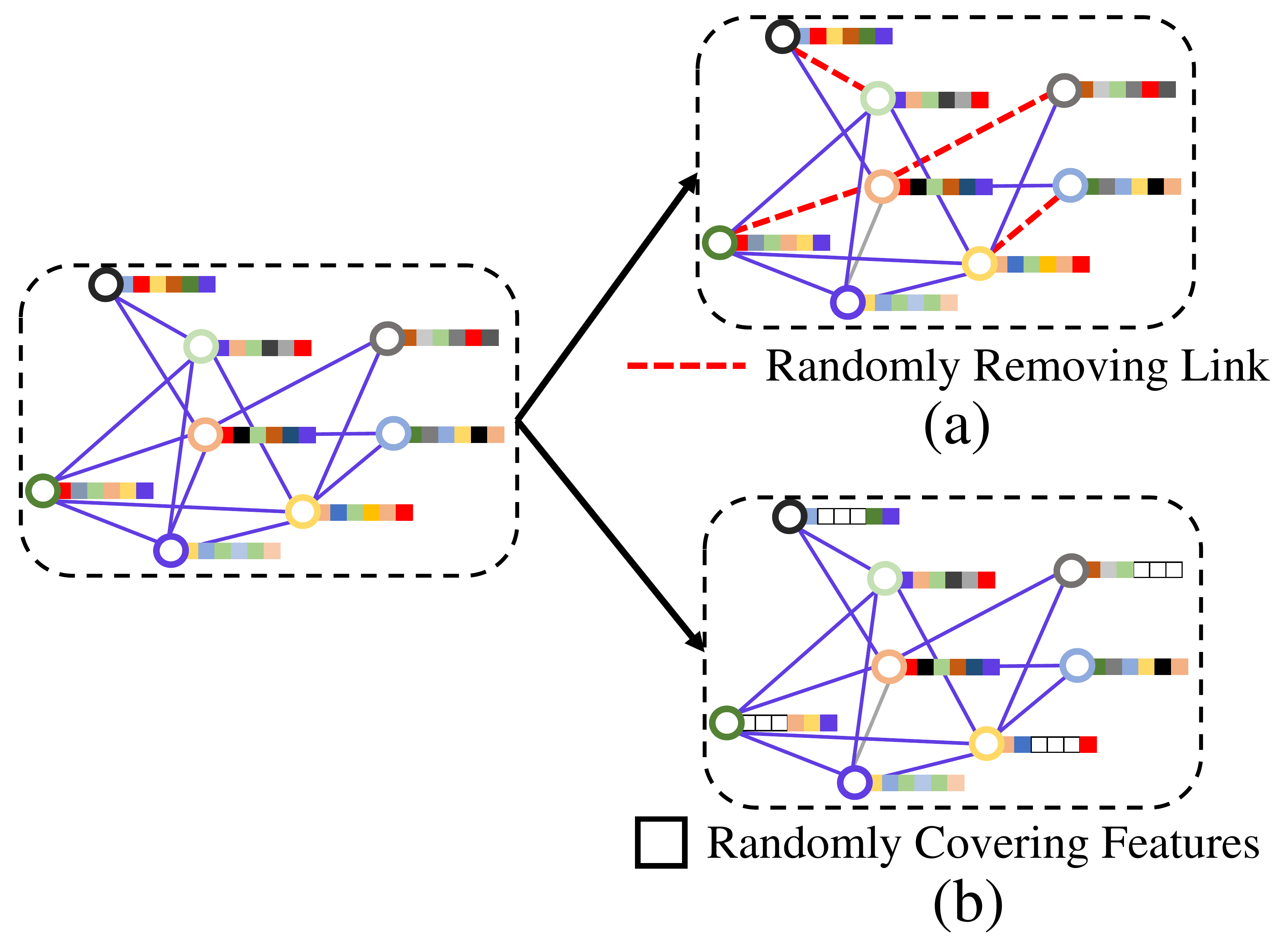}
  \caption{Illustration of the two types of self-supervised strategies, Randomly Removing Links (RRL) and Randomly Covering Features (RCF).}\label{SelfSup}
\end{figure}

\section{Method}
We first present the basic notations used in the paper.
Let $G({\cal V},{\cal E})$ represent an undirected graph with $\left|{\cal V}\right| = N$ nodes and edges $({v_i},{v_j}) \in {\cal E}$. Moreover, ${\mathop{\rm A}\nolimits}  \in {\mathbb{R}^{N \times N}}$ is the adjacency matrix and ${\mathop{\rm D}\nolimits}  \in {\mathbb{R}^{N \times N}}$ is the degree matrix of $G({\cal V},{\cal E})$, where ${{\mathop{\rm D}\nolimits} _{ii}} = \sum\nolimits_j {{{\mathop{\rm A}\nolimits} _{ij}}}$.  Accordingly, ${\mathop{\rm A}\nolimits} _{ij} = 1$ represents a edge existing between nodes ${v_i}$ and  ${v_j}$,  while each node  ${v_i} \in {\cal V}$ possesses a graph signal $x \in {\mathbb{R}^{N}}$, which is a feature vector.

\subsection{Learning Framework}
Figure~\ref{ProposedModel} presents an overview of the proposed framework, which consists of three parts: Input Module, Link Prediction Module and Objective Task Module.
In the SSL phase, the input data is modified by our proposed SSL, and the objective task is link prediction.
After SSL, the model convert to another task - classification. In this phase, the input data is without modified, and the objective task is classification which can classify each node ${v_i}$ into different categories based on node representation. More details of the proposed method are presented in the following paragraphs.

\subsubsection{Feature Learning:}
This module is the main part of the whole framework, which takes a graph with node features as input and produces a latent feature representation for each node. In the SSL phase, this module is trained jointly with the \textbf{Link prediction} module for the link prediction task; the input of this module is the graph data with incomplete information, such as missing part of edges or features. Through the self-supervised learning task of link prediction, the feature learning module explores both node representation and graph struct information during training.
In the classification phase, this module and the \textbf{Classification} module are trained jointly in an end-to-end manner for node classification. The architecture of this part consists of several graph convolutional layers; either spectral convolutional or spatial convolutional layers can be used in this module. Taking the most commonly used spectral convolutional layer, which is approximated by a first-order Chebyshev polynomial, as an example, the definition of a spectral convolutional layer is as follows:
\begin{align}\label{Eq0}
{H^{i + 1}} = {\tilde D^{ - \frac{1}{2}}}\tilde A{\tilde D^{ - \frac{1}{2}}}{H^{i}}\Theta
\end{align}
where $\Theta  \in { \mathbb{R}^ {C \times F}}$ is a matrix of filter parameters, ${H^{i+1}} \in {\mathbb{R}^{N \times F}}$  and ${H^{i}}$ are the output and input of the layer respectively, and $F$ denotes the number of filters, $\tilde A = {I_N} + {\mathop{\rm A}\nolimits}$ and ${\tilde D_{ii}} = \sum\nolimits_j {{{\tilde A}_{ij}}}$.
Similar to CNNs, each layer takes the output of the previous layer as input and learns the latent feature representation of each node.
The major difference is that each node inside graph can only learn the node representation from its neighborhoods.

\subsubsection{Classification:}
The task of this module is to classify node representation learned from the feature learning module into different categories in the classification phase. The classification module consists of several graph convolutional layers and a softmax function. This process can be formulated as follows:
\begin{align}\label{Eq1}
Z = softmax({H^i})
\end{align}
where ${H^i} = GCN(X,A)$ is the node representation learned by the GCN layers.

During training, the cross-entropy error over all labeled nodes are computed as the loss ${L_C}$, which is defined as follows:
\begin{align}\label{Eq2}
{L_C} =  - \sum\limits_{i \in {V_l}} {\sum\limits_{j = 1}^N {{M_{ij}}\log ({Z_{ij}})} }
\end{align}
where ${V_l}$ is the set of indices of labeled vertices, while $N$ is the dimension of the features, which is equal to the number of classes. Finally, ${M_{ij}}$ is the label indicator matrix.

\subsection{Link Prediction and Feature Completion}
Inspired by the advances made by SSL in improving the ability of feature learning, we also introduce SSL into GCN for improving the performance. Normally, the SSL process comprises the following steps. The first step involves digging the information from the image or video itself and designing a specific training strategy; the second involves training the SSL network, following the training strategy; the third involves using the weights provided by the SSL network to initialize the target network; finally, the fourth step fine-tunes the target network on the training dataset.
Following the above steps, we design two types of  SSL strategies, Randomly Removing Links (RRL) and Randomly Covering Features (RCF), to make the GCN learn the node representation from the graph data itself.  Both SSL processes are achieved by the link prediction module. This module, the objective of which is to predict whether a link (edge) exists between two nodes, can be formulated as follows:
\begin{align}\label{Eq3}
\hat A = sigmoid(H^i{(H^i)^T})
\end{align}
where ${H^i}$ is the node representation learned by the GCN layers, while $\hat A $ is the reconstructed adjacency matrix.

During training, since the adjacency matrix $A$ is a sparse matrix, the weighted cross-entropy ${L_L}$ is used as the loss function, which is defined as follows:
\begin{align}\label{Eq4}
L_{L}  =  - \sum\limits_{i} W({A_{i}\log ( \hat A_{i})) + (1 - A_{i})\log (1 - \hat A_{i})}
\end{align}
where ${i}$ is the set of indices of the adjacency matrix and $W$ is the positive weight, which is equal to the ratio of the number of non-links to links in adjacency matrix $A$.


\subsubsection{Randomly Removing Links (RRL)}
The graph edge, which is one of the most important structures in graph data, stores the graph structure information and represents the relationship between nodes. For example, in chemoinformatics, chemical compounds are often represented as molecular graphs in which the nodes correspond to their atoms and the edges correspond to the chemical bonds among them. In citation networks, moreover, the graph node represents a document and the graph edge represents the existence of a citation link  between them. In addition, the GCN also learns the node representation along the edge. To make the GCN able to utilize the structure information of the graph, we design a SSL strategy called Randomly Removing Links (RRL) for SSL, as shown in Figure~\ref{SelfSup}(a). This pretext task first randomly removes parts of the edges inside the graph, then requires the GCN model to predict the graph link of the input graph.
\begin{table}[t]
	\caption{Summary statistics of the citation network datasets.}
	\centering
	\begin{tabular}{l|c|c|c}
		\hline
		\textbf{Dataset}  & Citeseer & Cora & Pubmed \\ \hline
		\# Nodes            & 3,327  & 2,708  & 19,717 \\\hline
		\# Edges            & 4,732  & 5,429  & 44,338 \\ \hline
		\# Features         & 3,327  & 1,433  & 500  \\ \hline
		\# Categories       & 6  & 7  & 3  \\ \hline
		\# Training Nodes   & 120  & 140  & 60 \\ \hline
		\# Validation Nodes & 500  & 500  & 500 \\ \hline
		\# Test Nodes       & 1000  & 1000  & 1000 \\ \hline
	\end{tabular}
	\label{Tabel_citation}
\end{table}
\begin{table}
\caption{Exploring the quality of the self-supervised learned features w.r.t. the percentage of randomly removing links and covering features.
The performance of GCN on the Cora dataset with varying categories.}
\centering
\begin{tabular}{c|c|c|c}
\hline
\hline
\multicolumn{2}{l|}{GCN}                              & ACC{[}\%{]} & P-Value \\ \hline
\multicolumn{2}{l|}{Without SSL}                          & 81.51       & ---       \\ \hline
\multirow{3}{*}{10\%}                       & RRL      & 82.60       & 2.9E-4  \\ \cline{2-4}
                                            & RCF      & 82.47       & 2.3E-4  \\ \cline{2-4}
                                            & RRL\&RCF & 83.73       & 1.6E-6  \\ \hline
\multicolumn{1}{c|}{\multirow{3}{*}{20\%}} & RRL      & 82.83       & 1.3E-7  \\ \cline{2-4}
\multicolumn{1}{c|}{}                      & RCF      & 82.26       & 5.8E-4  \\ \cline{2-4}
\multicolumn{1}{c|}{}                      & RRL\&RCF & 83.59       & 1.1E-6  \\ \hline
\multirow{3}{*}{30\%}                       & RRL      & 82.71       & 8.1E-5  \\ \cline{2-4}
                                            & RCF      & 82.44       & 3.7E-4  \\ \cline{2-4}
                                            & RRL\&RCF & 83.61       & 1.7E-5  \\ \hline
\multirow{3}{*}{40\%}                       & RRL      & 83.60       & 6.1E-7  \\ \cline{2-4}
                                            & RCF      & 82.65       & 2.1E-6  \\ \cline{2-4}
                                            & RRL\&RCF & 83.80       & 1.4E-5  \\ \hline
\multirow{3}{*}{50\%}                       & RRL      & 82.58       & 3.8E-5  \\ \cline{2-4}
                                            & RCF      & 82.47       & 2.3E-4  \\ \cline{2-4}
                                            & RRL\&RCF & 83.52       & 4.1E-5  \\ \hline
\multirow{3}{*}{60\%}                       & RRL      & 82.43       & 8.6E-3  \\ \cline{2-4}
                                            & RCF      & 82.19       & 6.7E-4  \\ \cline{2-4}
                                            & RRL\&RCF & 83.70       & 1.3E-6  \\ \hline\hline
\end{tabular}
\label{Tabel_cora_experiment}
\end{table}

\subsubsection{Randomly Covering Features (RCF)}
Aside from the structure information, each node contains a feature vector that stores the node representation. The node with feature vector can be made to be completely analogous with the image, since the image information is stored inside the pixels and the feature vector contains all feature information of each node in the graph. In the image domain, image inpainting~\cite{pathak2016context} is an effective pretext task for ensuring that the model understands the content of an image. This pretext task requires the model to reconstruct the missing parts that are erased randomly during training. Many experiments have proven that this pretext task can significantly improve the model performance. Inspired by the advances made in the image inpainting field, in this paper, we propose a similar self-supervised strategy, called Randomly Covering Features (RCF), to further exploit the information from the graph data, as shown in Figure~\ref{SelfSup}(b). In this pretext task, we randomly cover parts of the features inside the feature vector during training and supervise the model to predict the links between nodes.

\begin{figure}[t]
	\centering
	\includegraphics[width=0.9\columnwidth]{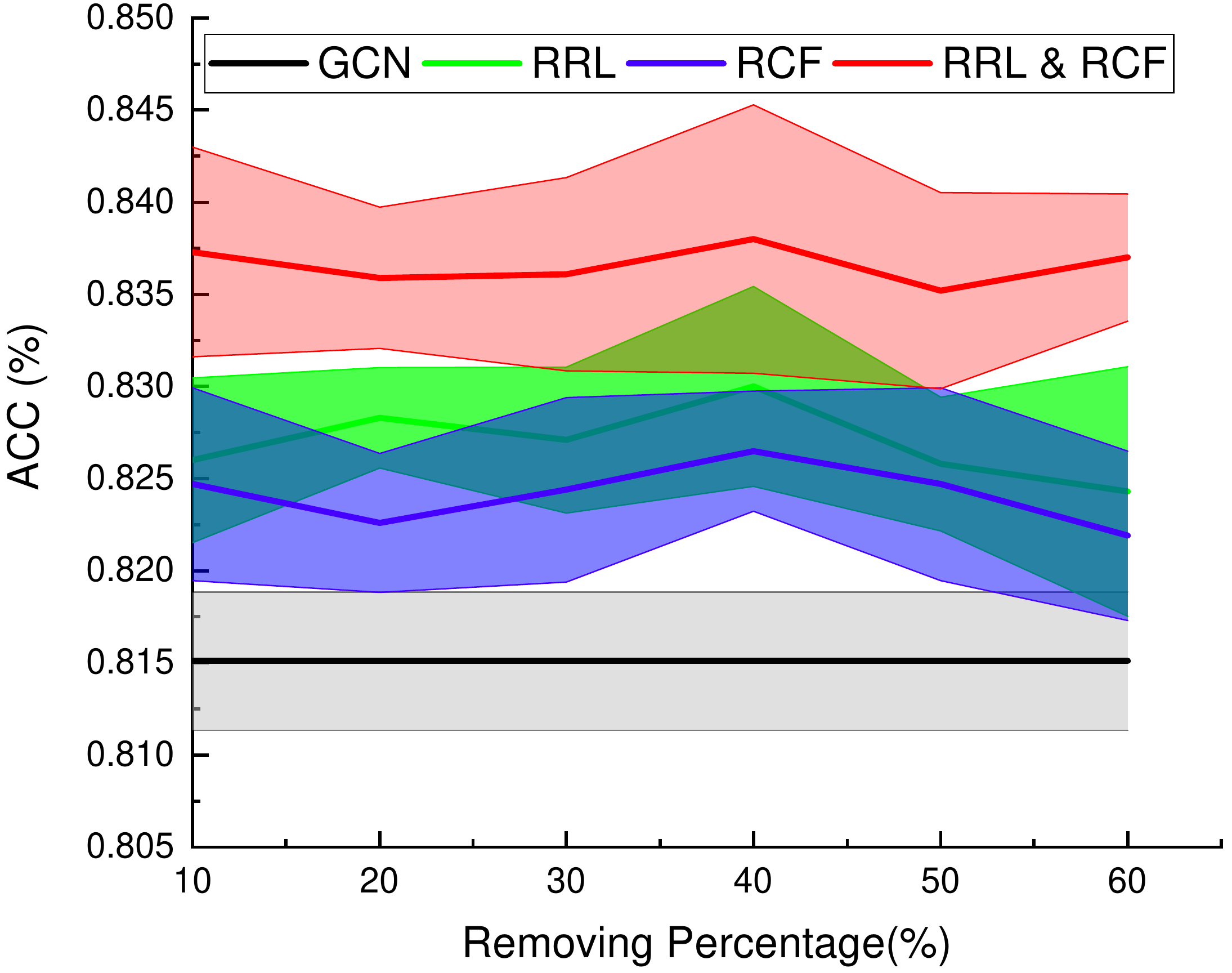}
	\caption{Classification performance of GCN on the Cora dataset with three types of self-supervised strategies and various percentages of removing and covering.}
	\label{figure_cora_experiment}
\end{figure}

\section{Experiments}
\begin{figure}
	\centering
	\subfigure[Input]{\includegraphics[width=0.3\columnwidth]{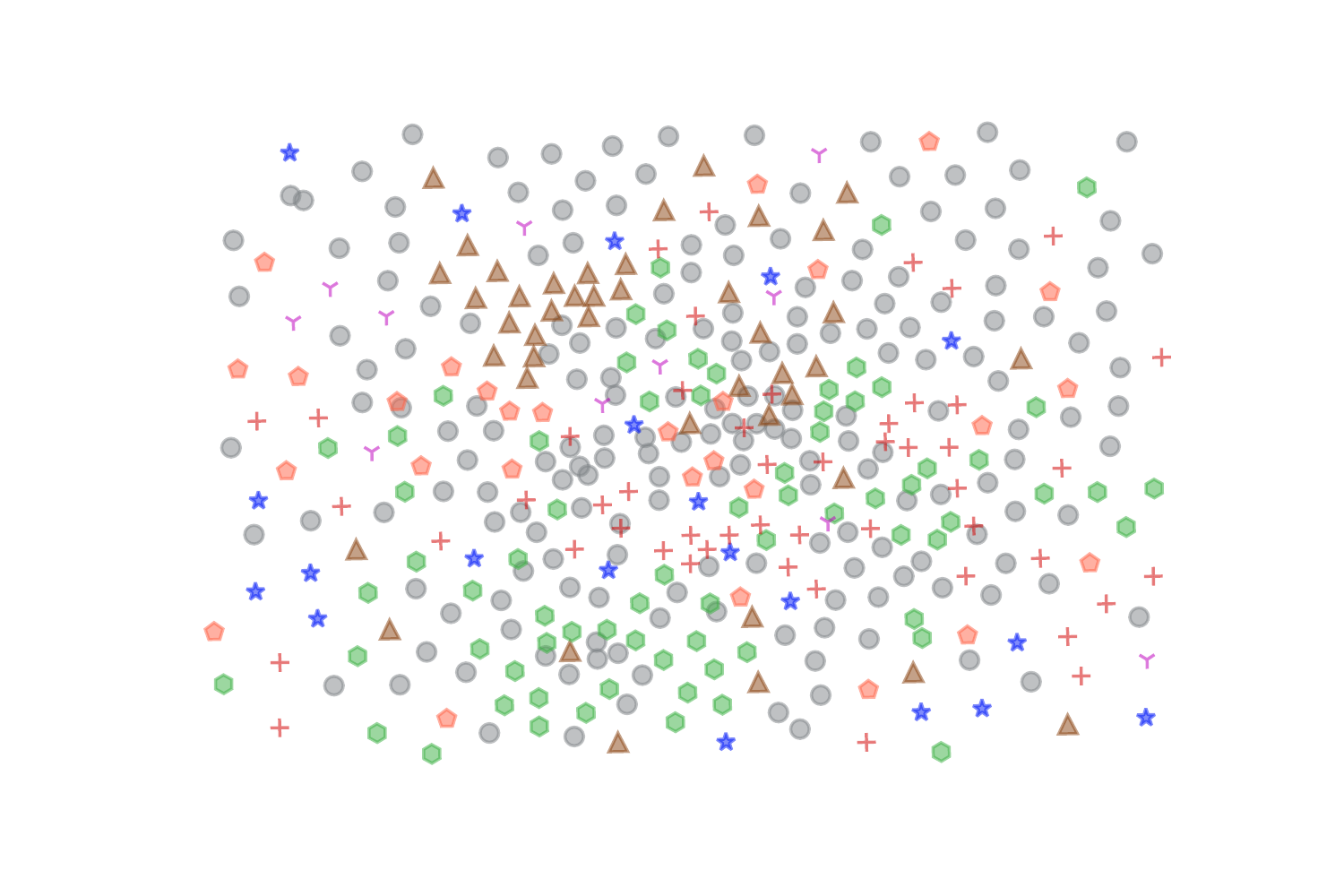}}
	\\
	\subfigure[RRL(10\%)]{\includegraphics[width=0.3\columnwidth]{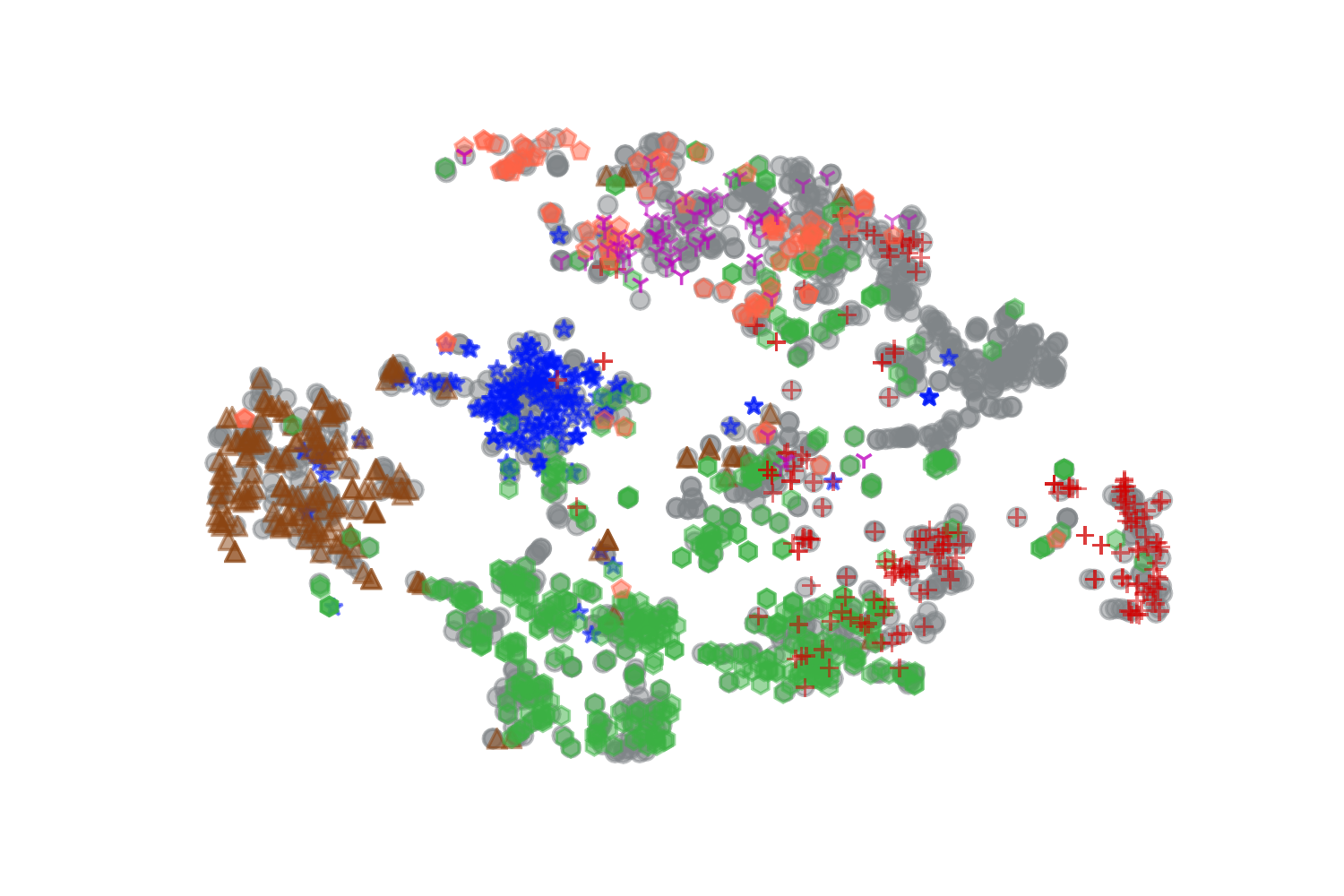}}
	\hspace{1mm}
	\subfigure[RCF(10\%)]{\includegraphics[width=0.3\columnwidth]{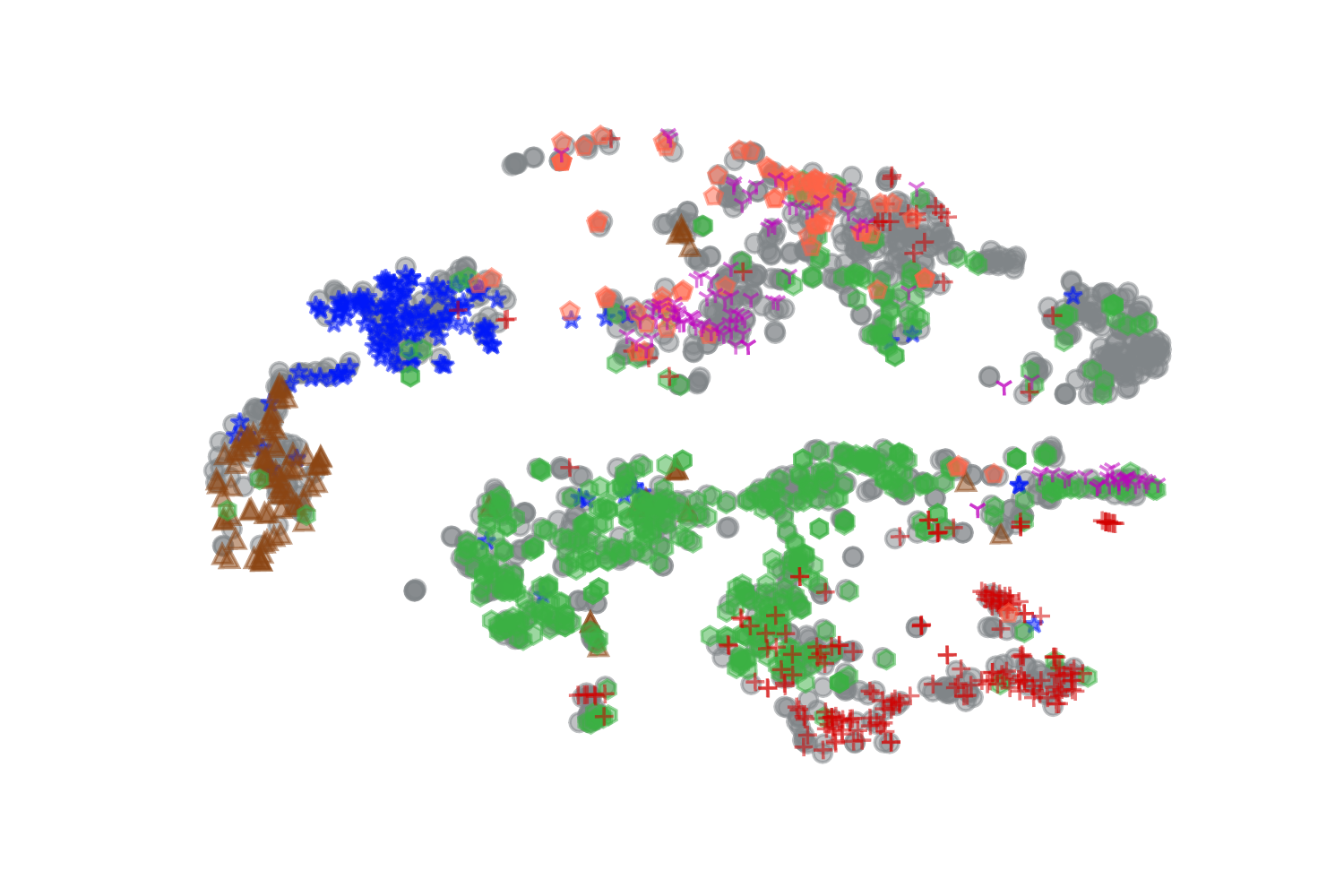}}
	\hspace{1mm}
	\subfigure[Mixed(10\%)]{\includegraphics[width=0.3\columnwidth]{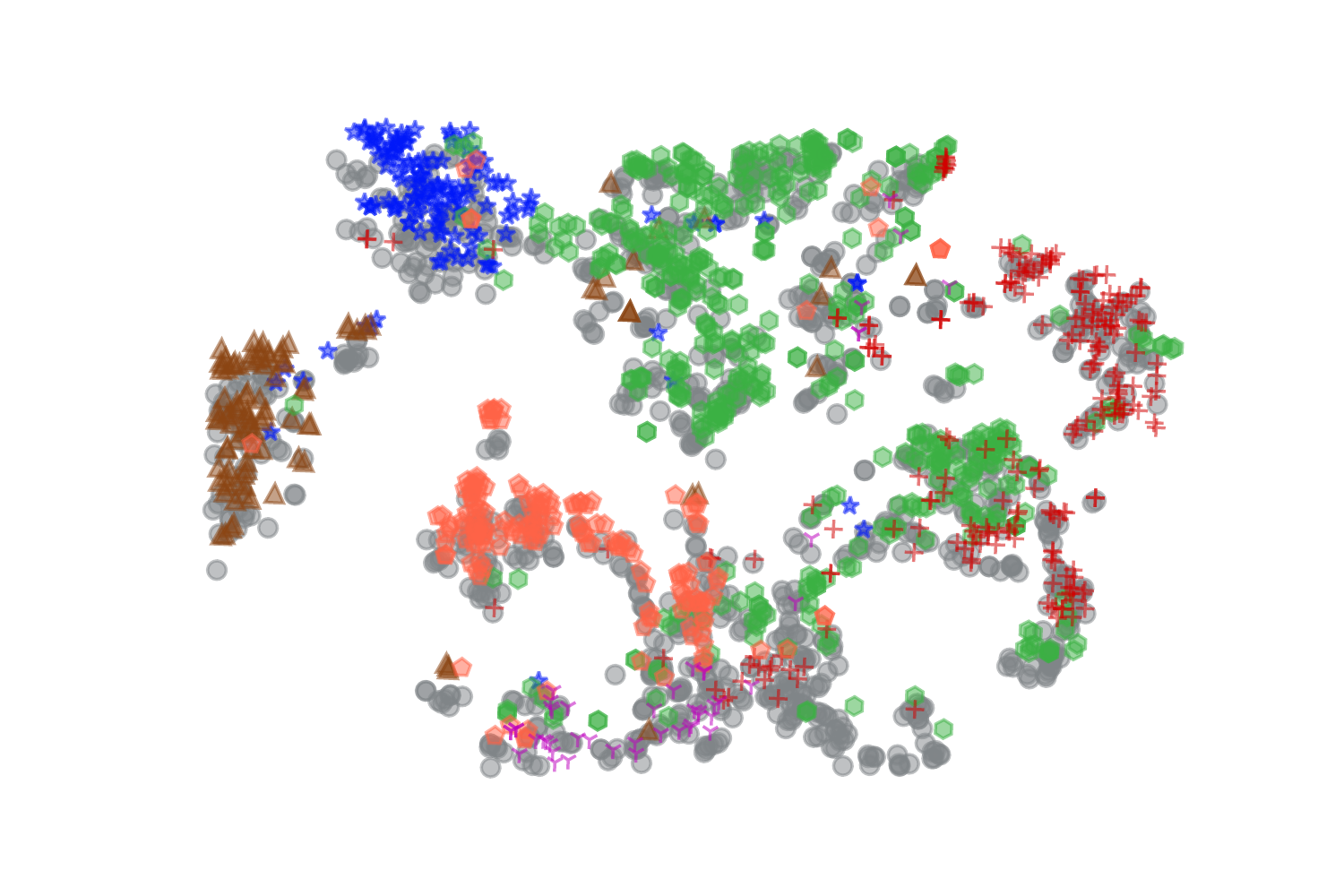}}
    \\

	\subfigure[RRL(20\%)]{\includegraphics[width=0.3\columnwidth]{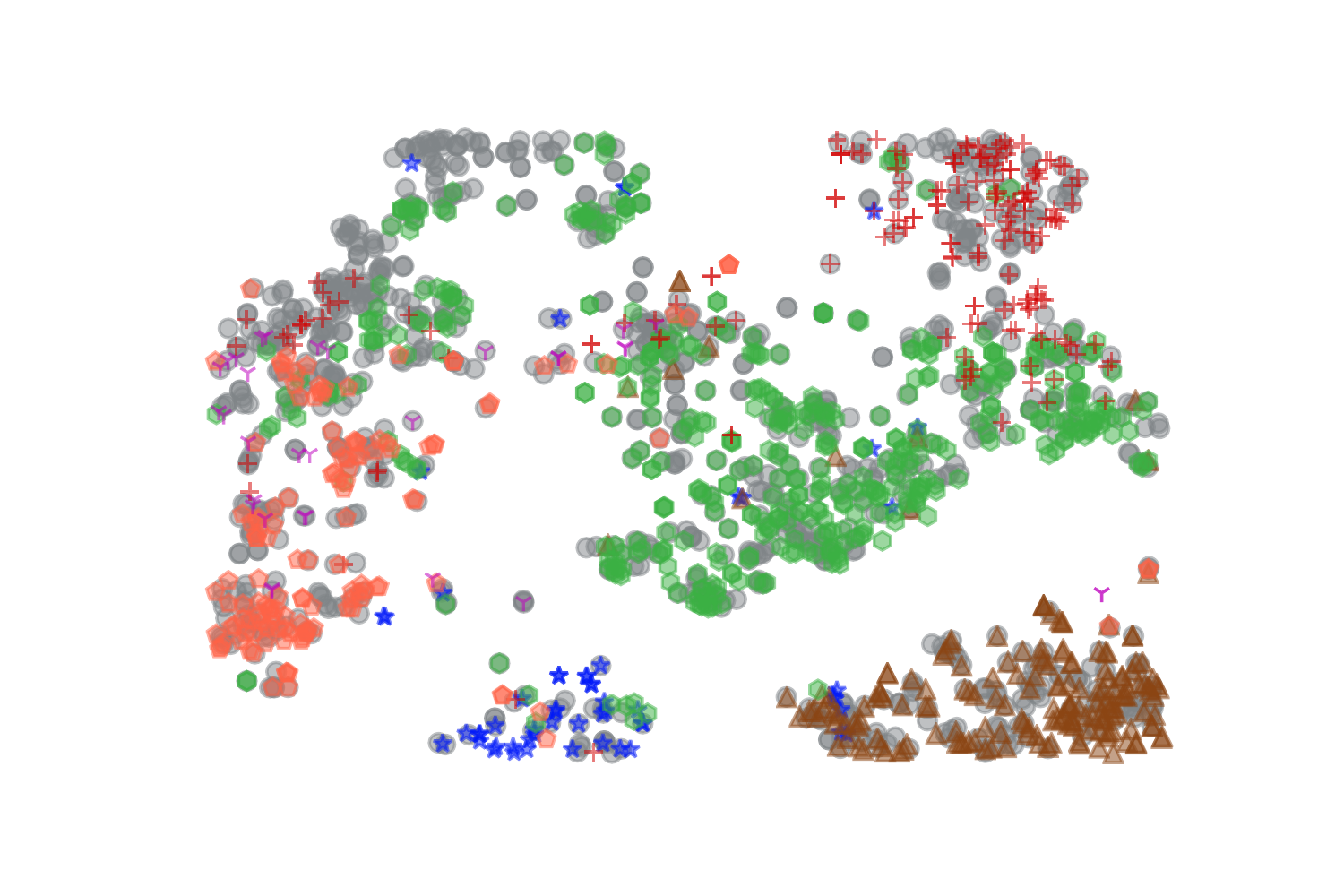}}
	\hspace{1mm}
	\subfigure[RCF(20\%)]{\includegraphics[width=0.3\columnwidth]{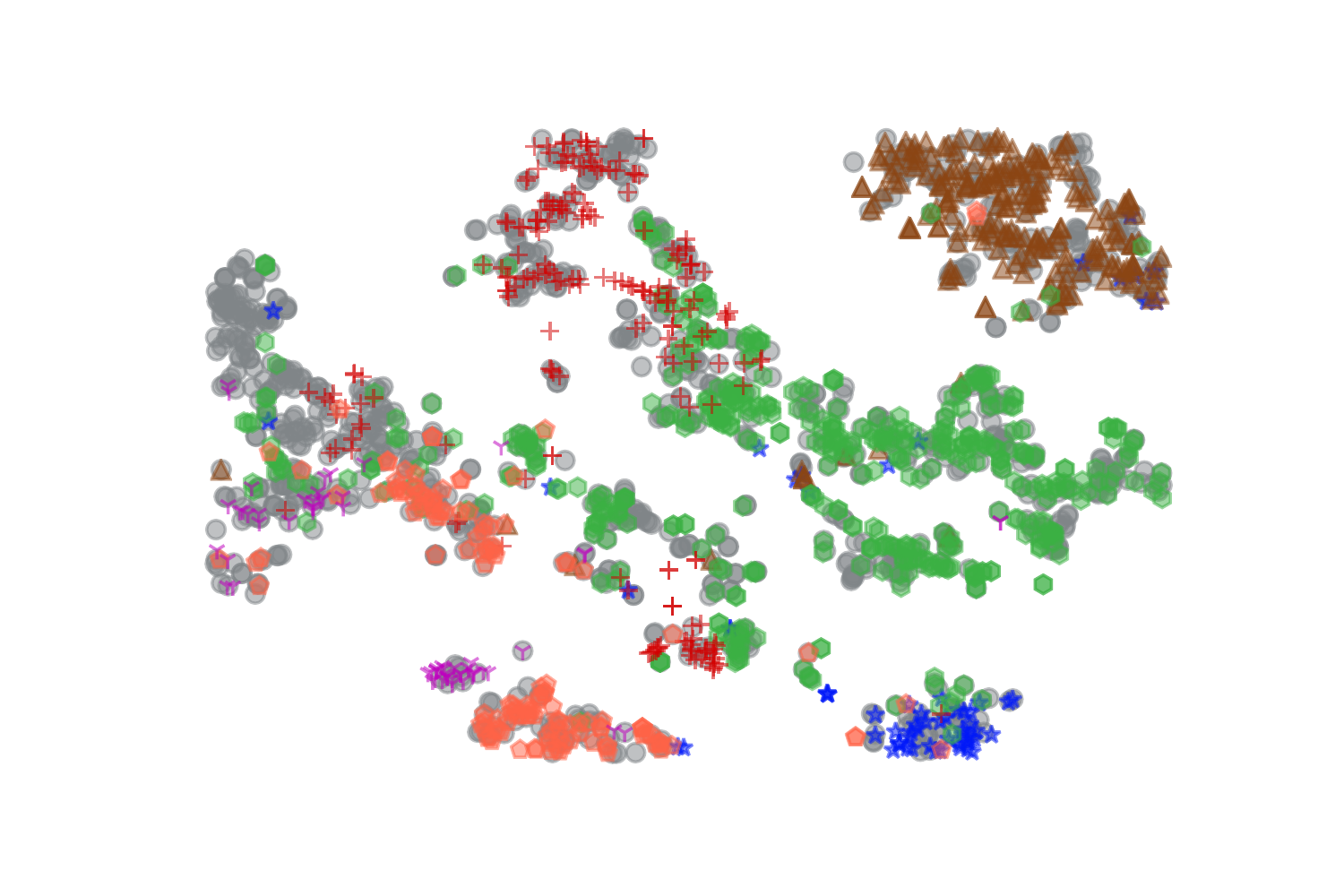}}
	\hspace{1mm}
	\subfigure[Mixed(20\%)]{\includegraphics[width=0.3\columnwidth]{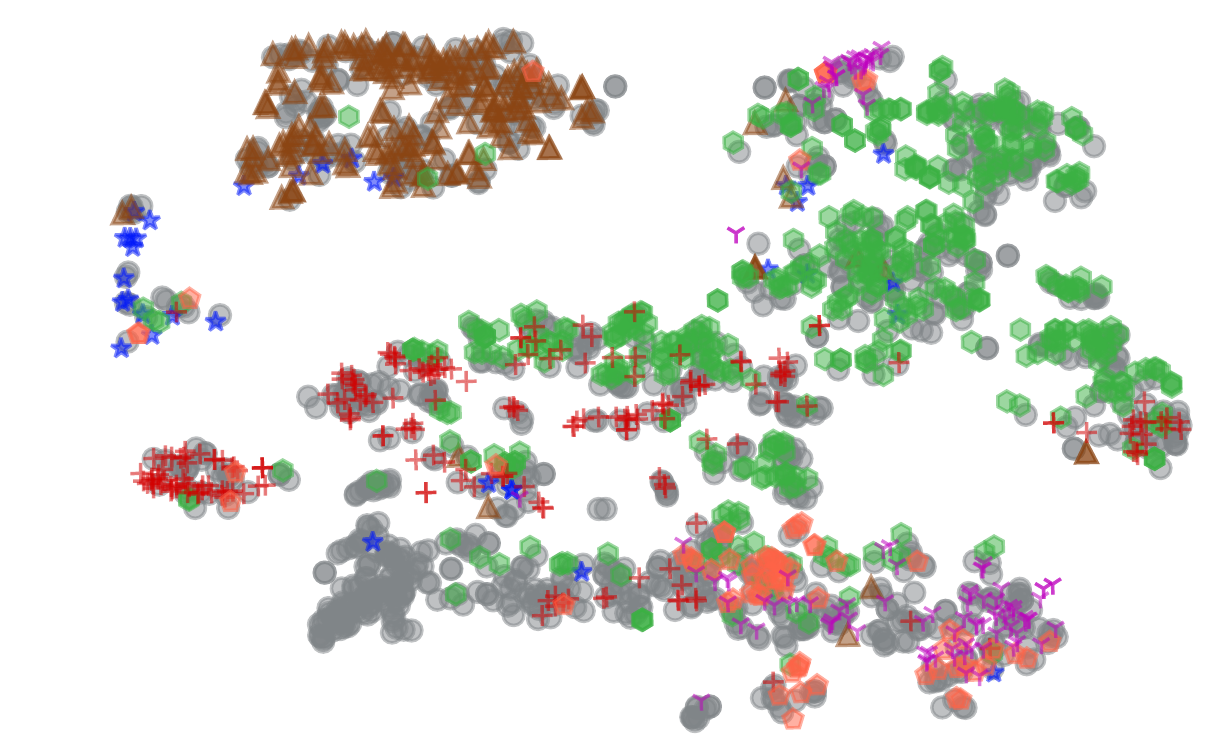}}
    \\

	\subfigure[RRL(30\%)]{\includegraphics[width=0.3\columnwidth]{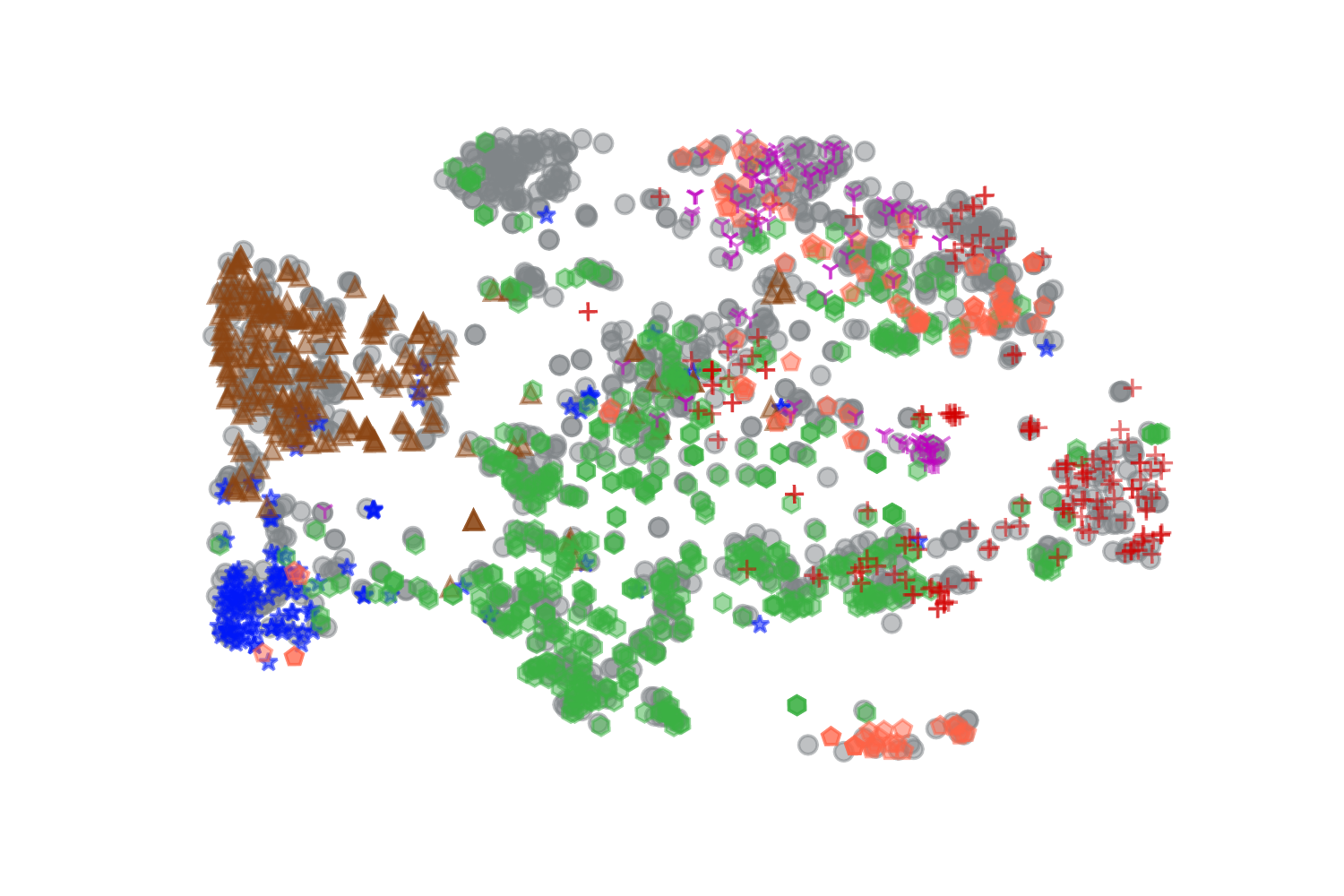}}
	\hspace{1mm}
	\subfigure[RCF(30\%)]{\includegraphics[width=0.3\columnwidth]{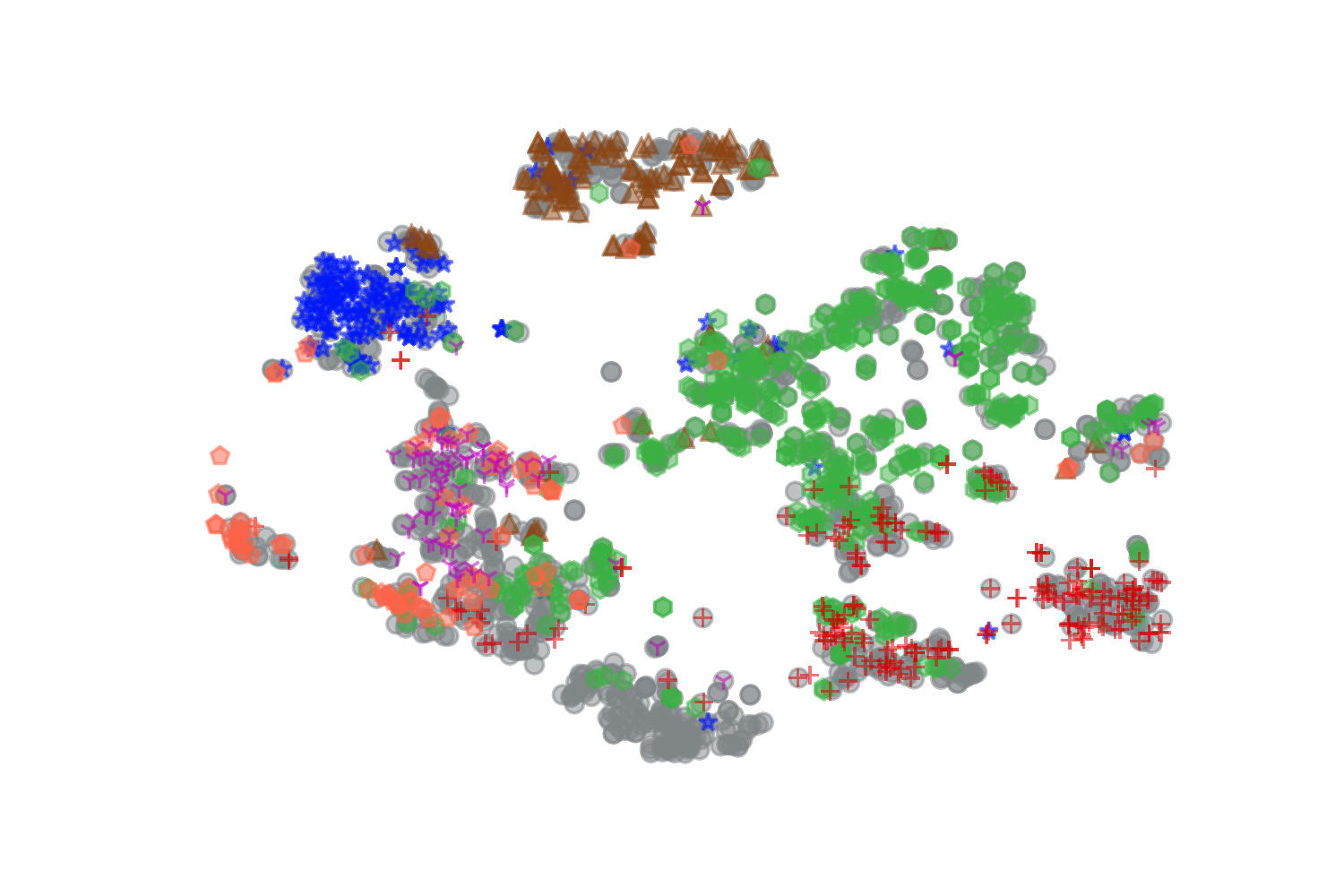}}
	\hspace{1mm}
	\subfigure[Mixed(30\%)]{\includegraphics[width=0.3\columnwidth]{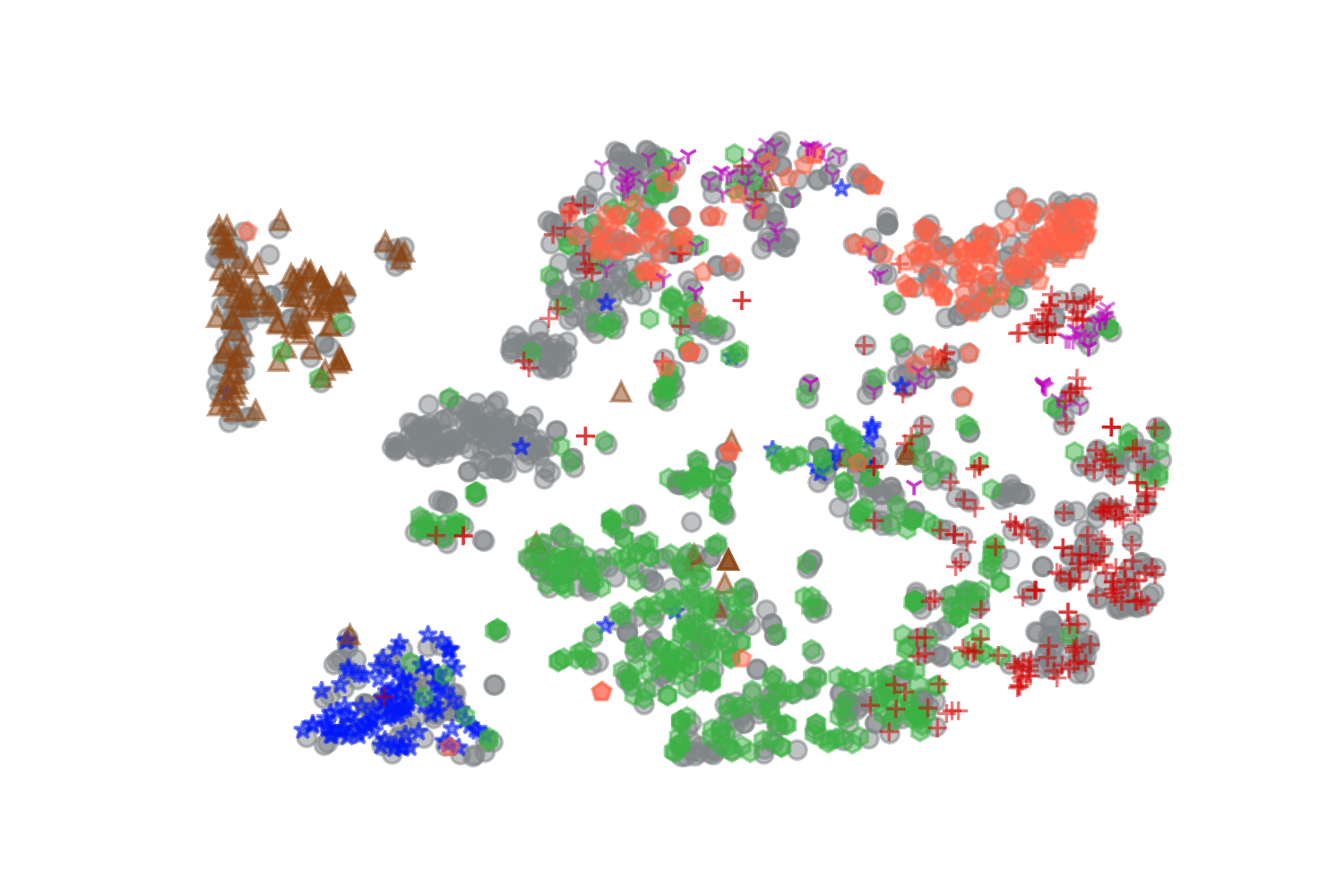}}
    \\
	\subfigure[RRL(40\%)]{\includegraphics[width=0.3\columnwidth]{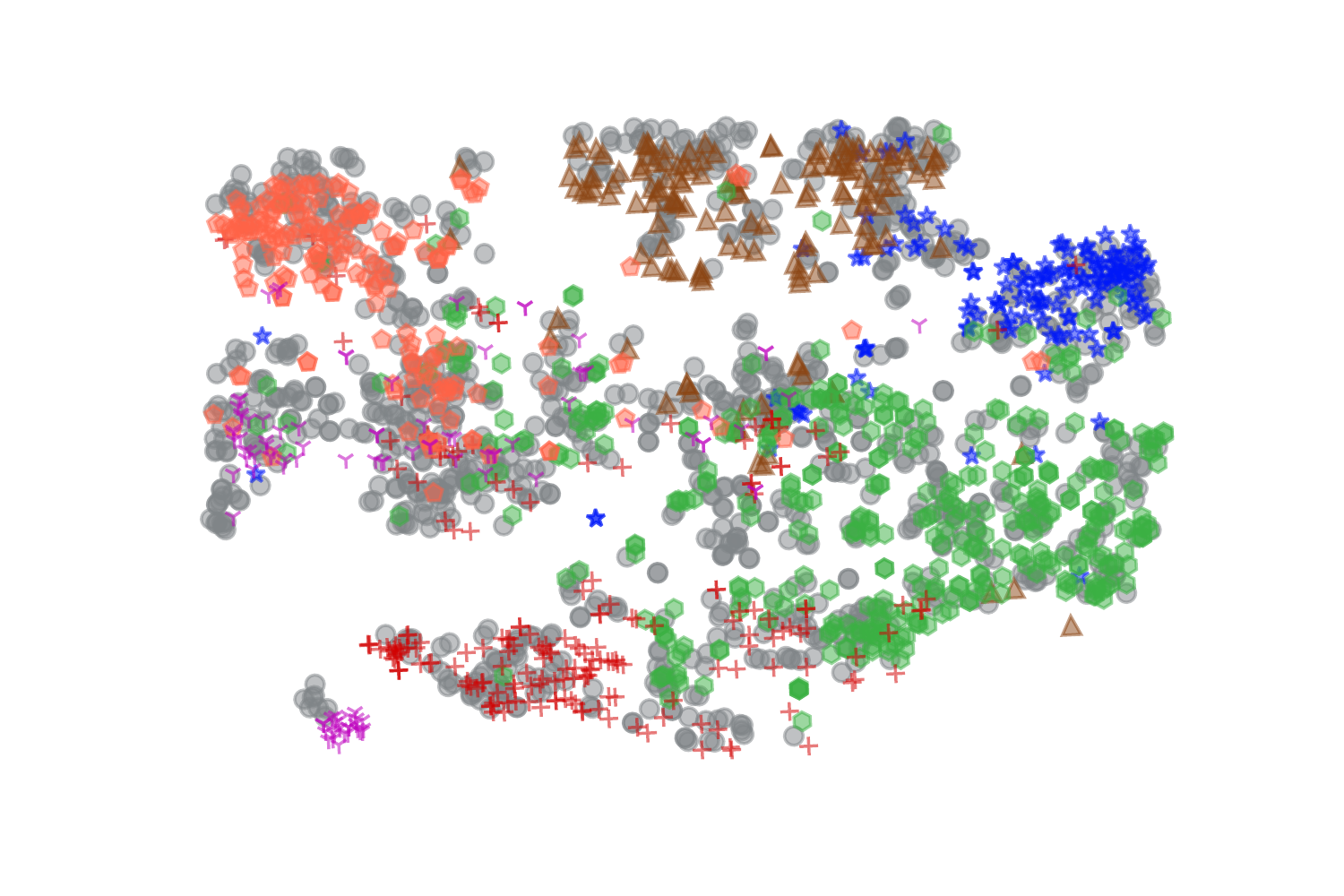}}
	\hspace{1mm}
	\subfigure[RCF(40\%)]{\includegraphics[width=0.3\columnwidth]{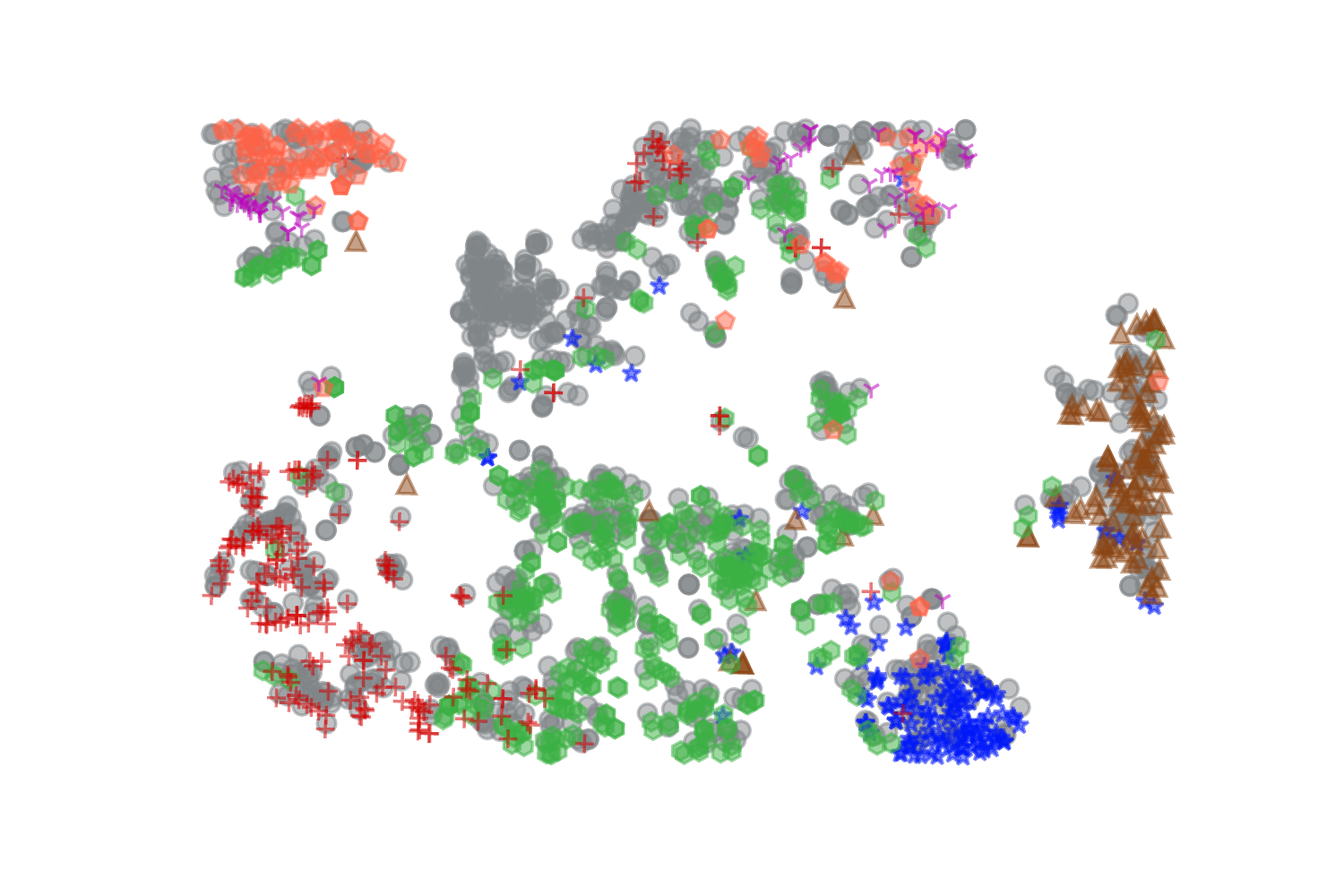}}
	\hspace{1mm}
	\subfigure[Mixed(40\%)]{\includegraphics[width=0.3\columnwidth]{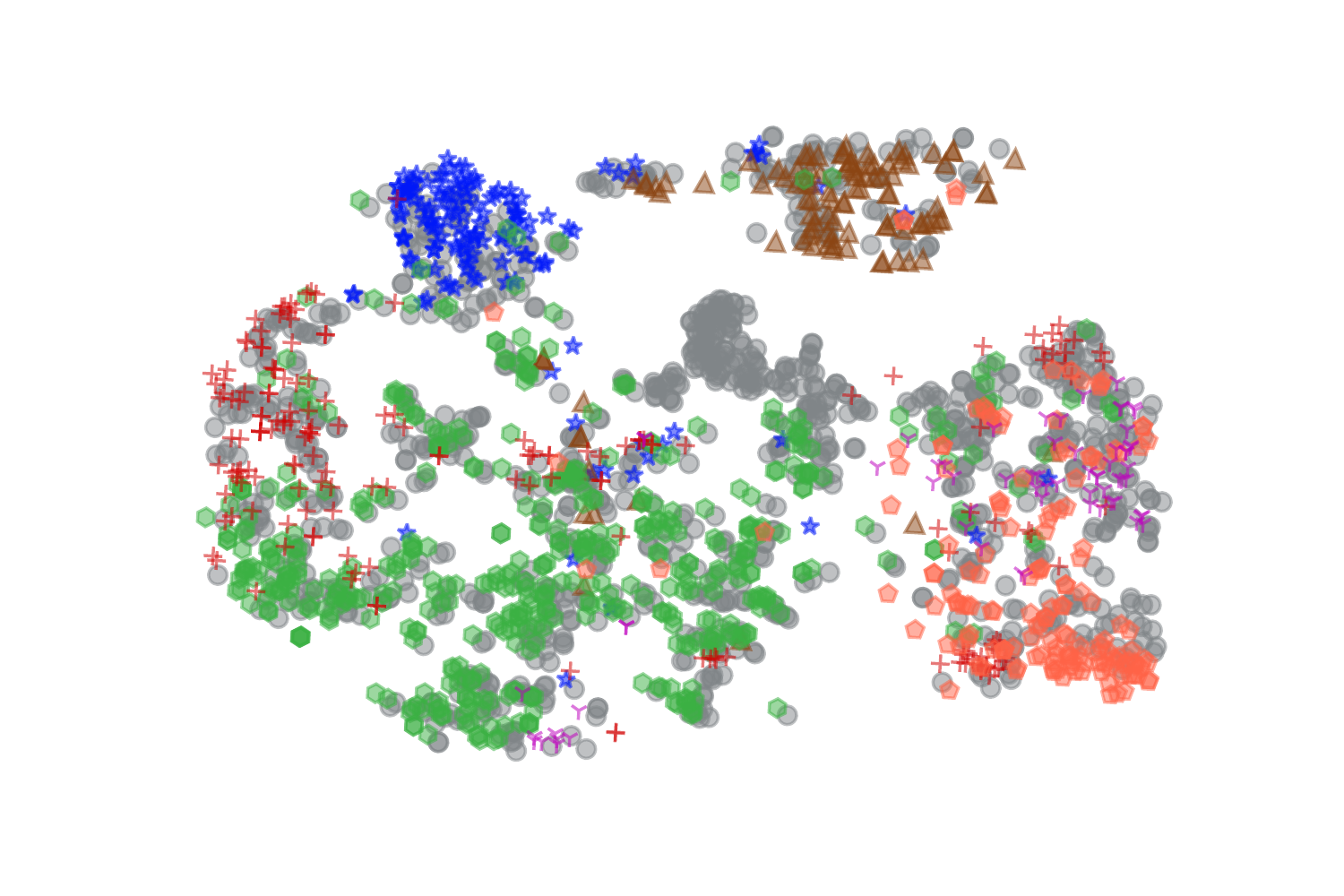}}

	\subfigure[RRL(50\%)]{\includegraphics[width=0.3\columnwidth]{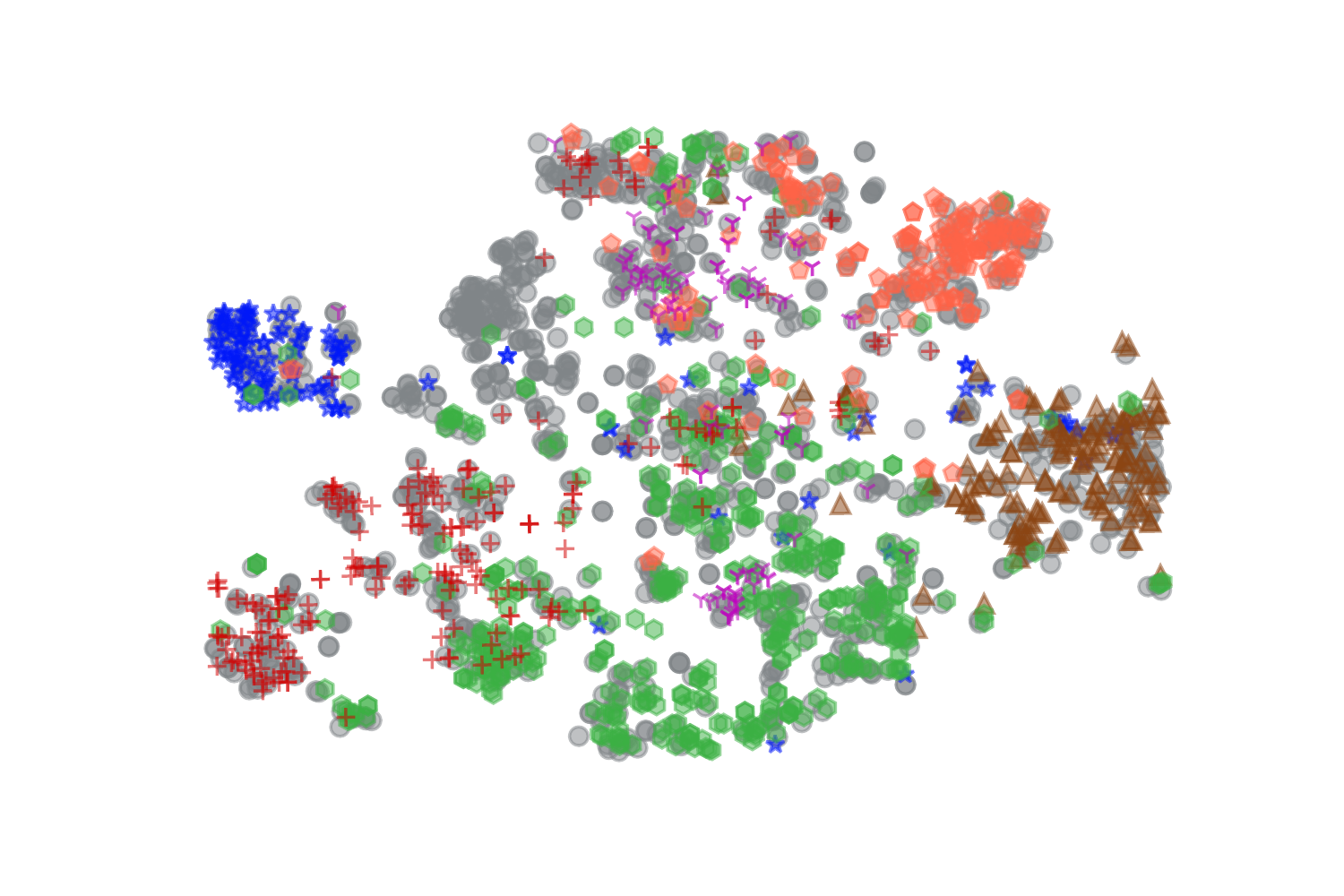}}
	\hspace{1mm}
	\subfigure[RCF(50\%)]{\includegraphics[width=0.3\columnwidth]{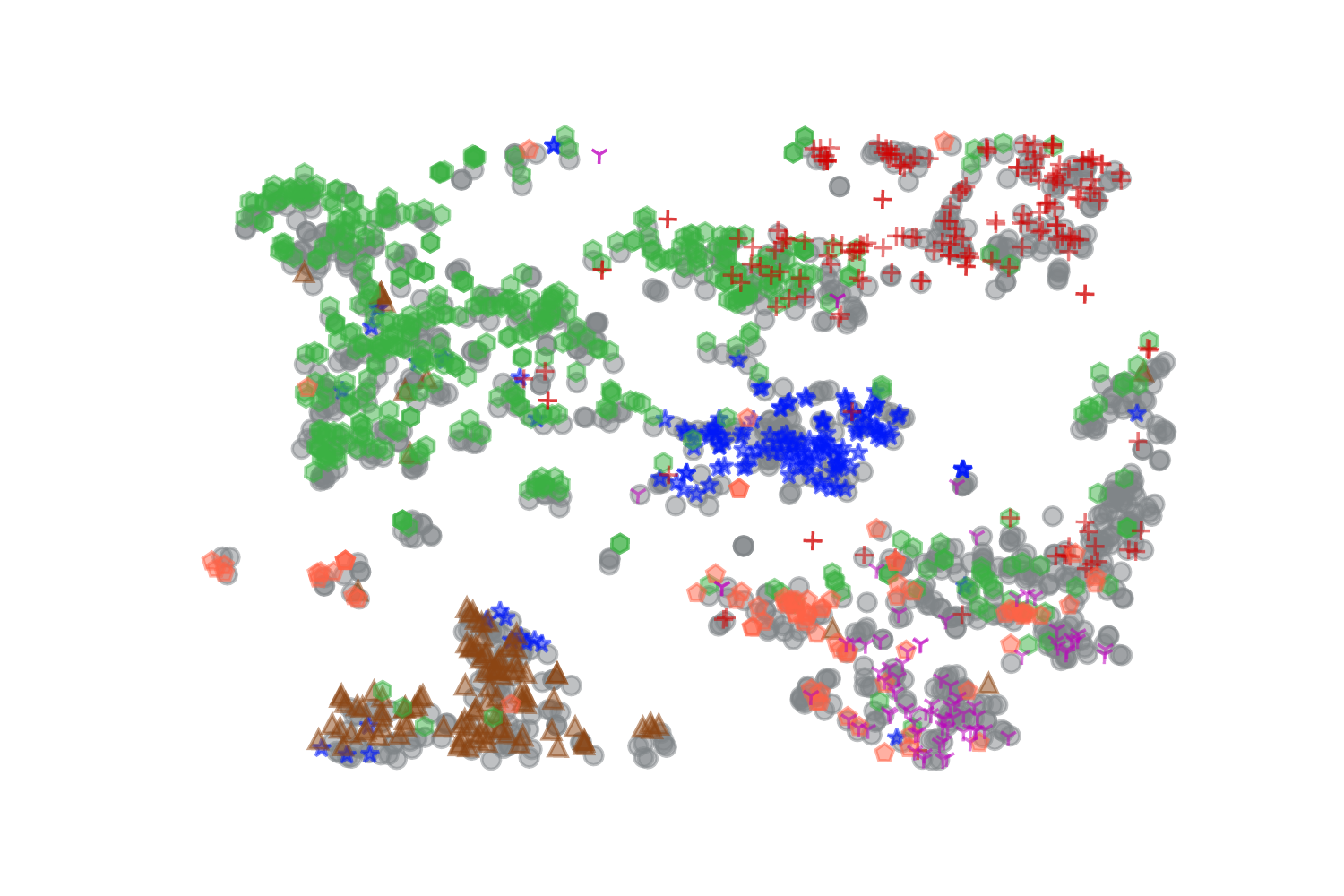}}
	\hspace{1mm}
	\subfigure[Mixed(50\%)]{\includegraphics[width=0.3\columnwidth]{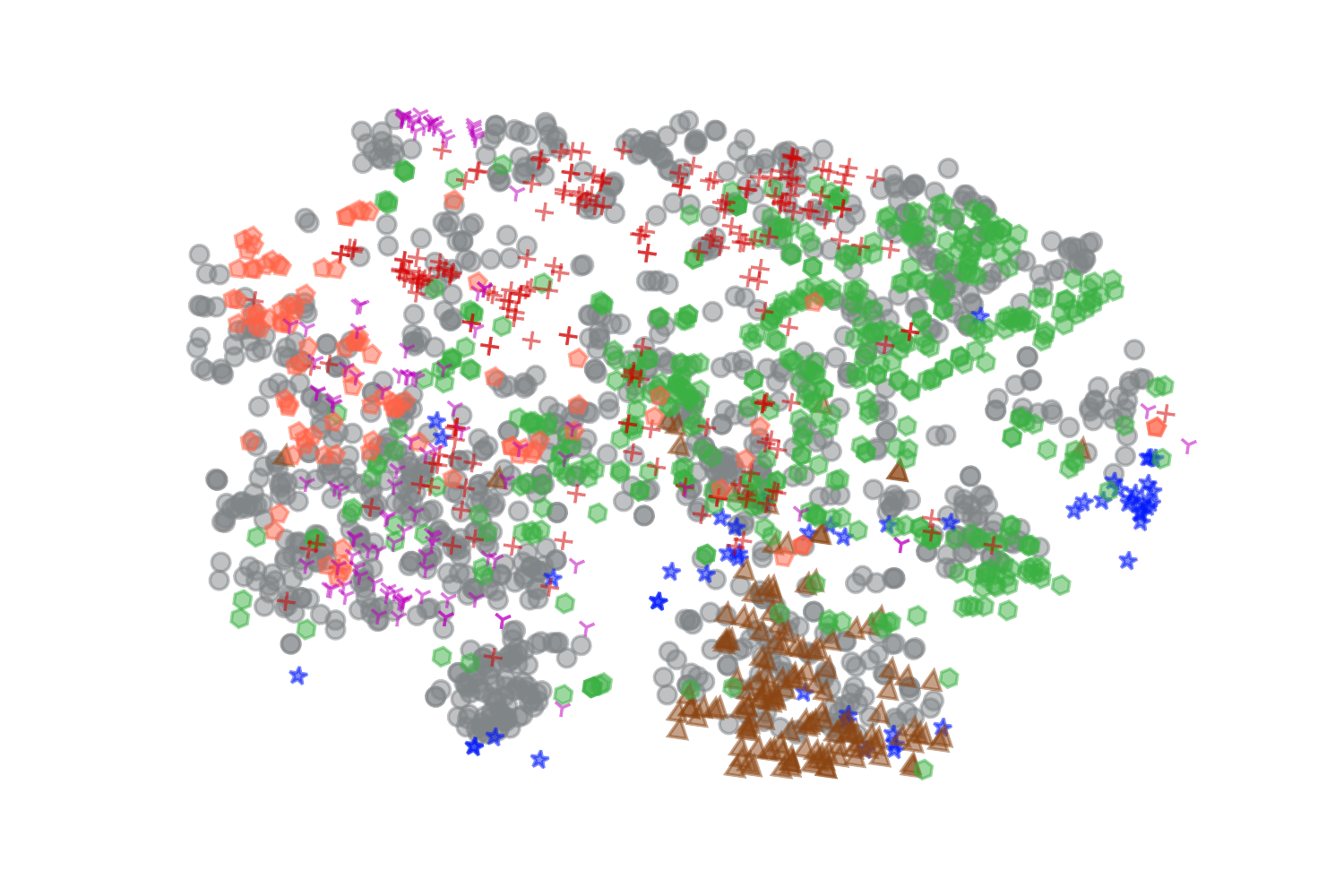}}
    \\
	\subfigure[RRL(60\%)]{\includegraphics[width=0.3\columnwidth]{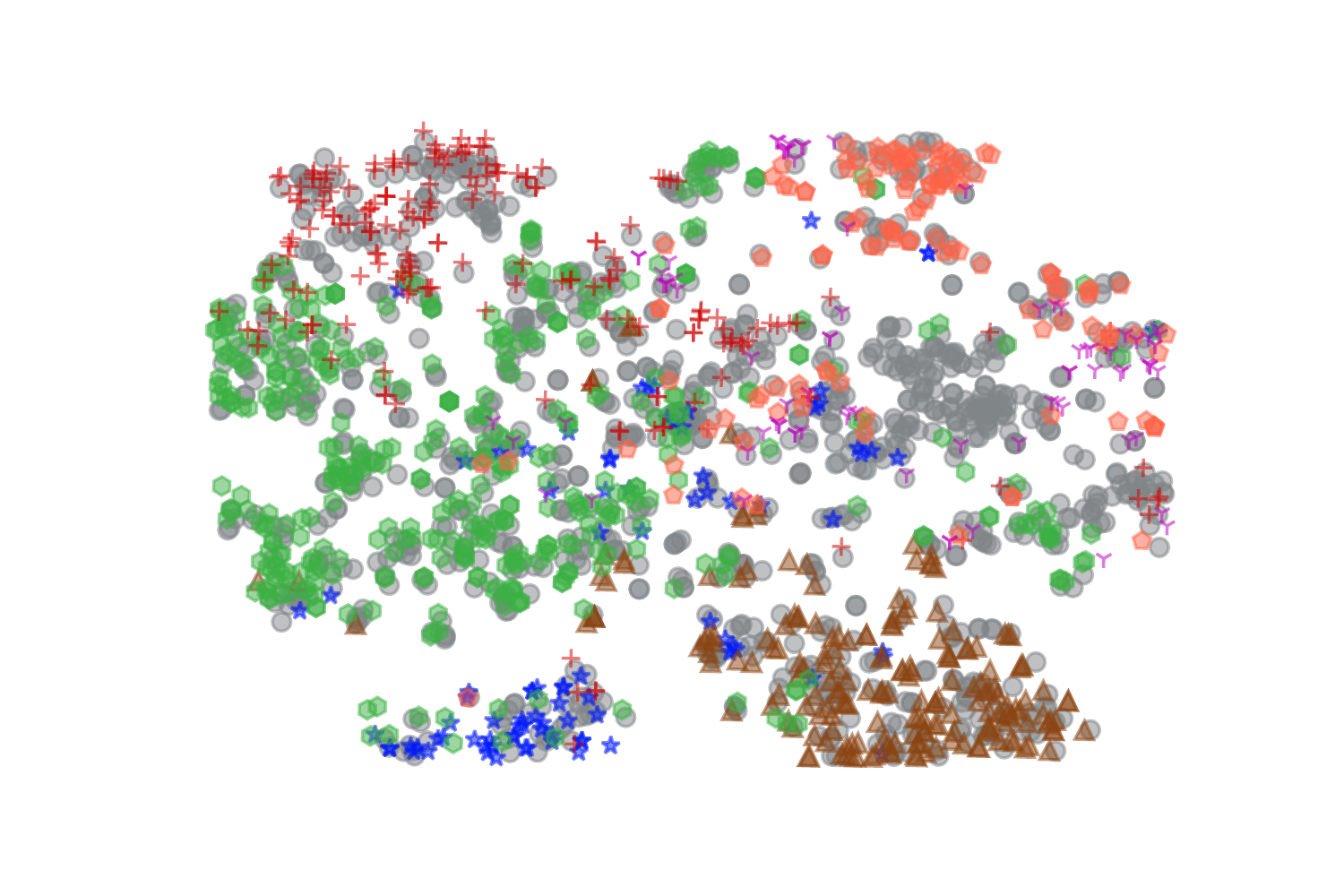}}
	\hspace{1mm}
	\subfigure[RCF(60\%)]{\includegraphics[width=0.3\columnwidth]{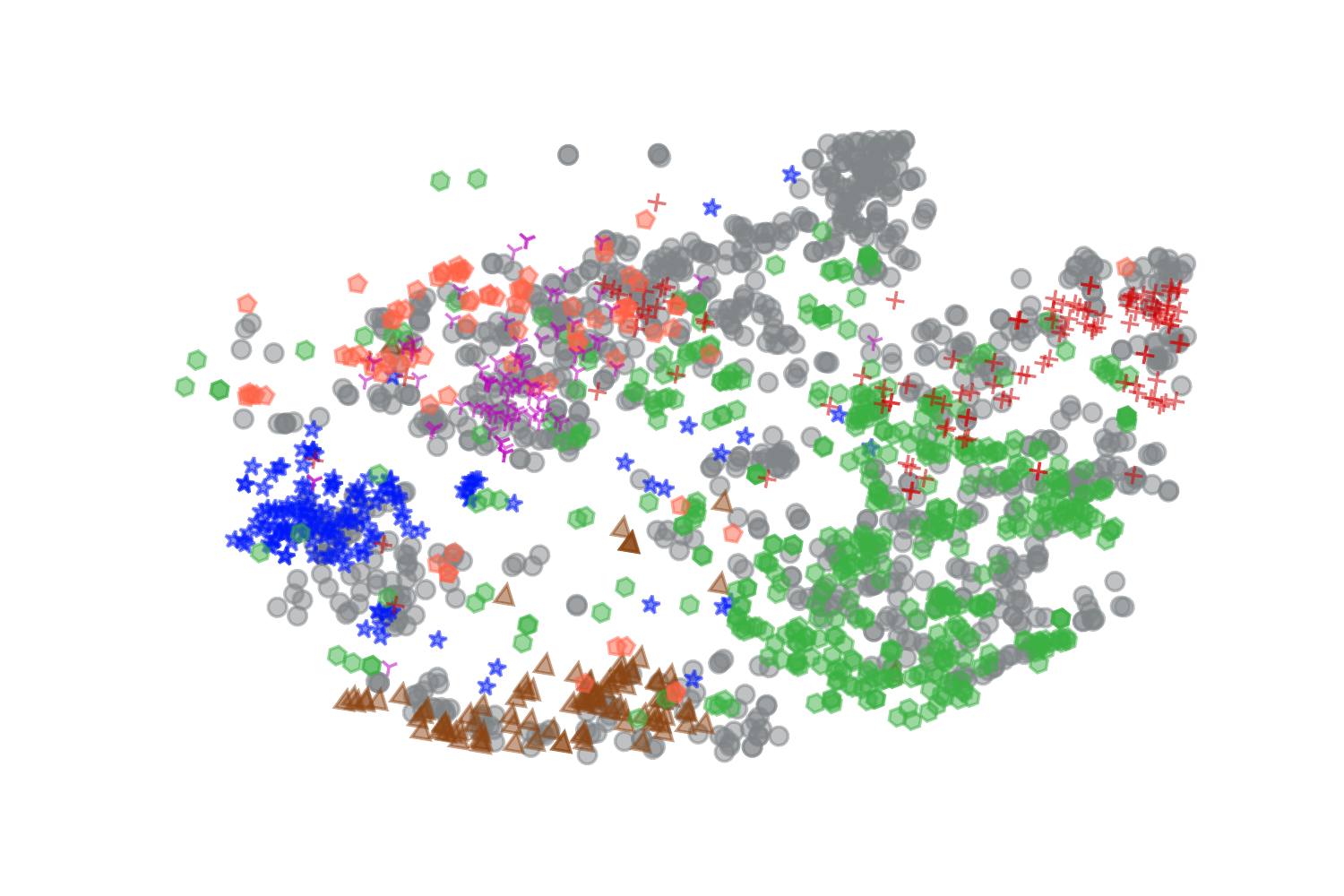}}
	\hspace{1mm}
	\subfigure[Mixed(60\%)]{\includegraphics[width=0.3\columnwidth]{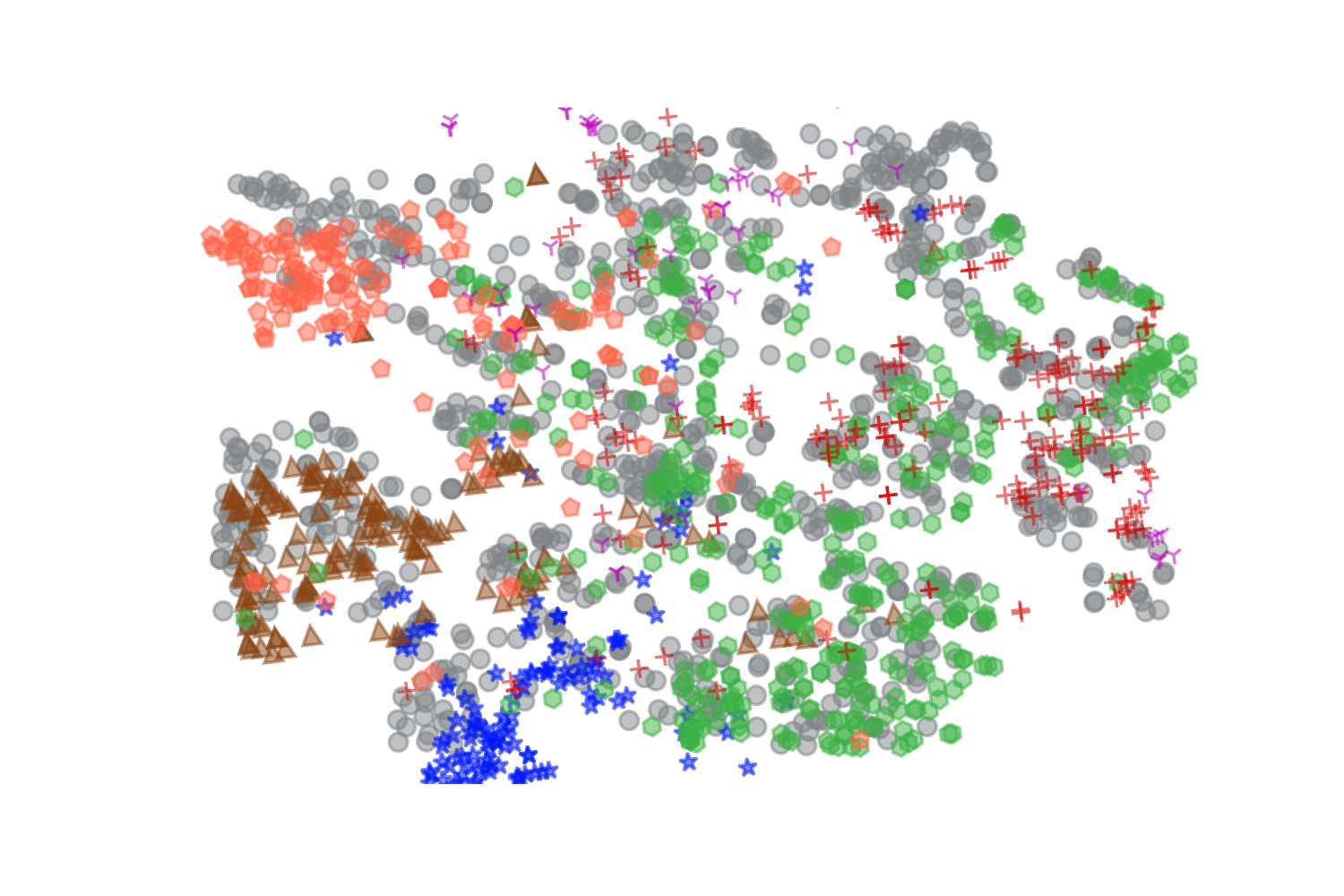}}
\caption{t-SNE visualization of the learned node representation using our proposed SSL method. ``Mixed'' represents the combined use of RRL and RCF. Different colors represent different categories.  }\label{TSNE}
\end{figure}
In this section, we first discuss the effects of SSL in improving the model performance, and then we apply our SSL on two representative GCN models - GCN~\cite{kipf2016semi}, GAT~\cite{velivckovic2017graph} - to demonstrate generalization as well as portability. Graph data from three publicly available citation network datasets, Citeseer, Cora and Pubmed, are used in our experiments. Details of the data and the implementation of our algorithm are presented as follows.
\subsection{Datasets}
Each of the three citation network datasets, Citeseer, Cora and Pubmed, contains a bag-of-words representation of the document and citation links between the documents. Following the experimental setup in \cite{yang2016revisiting}, we treat the bag-of-words as the feature vectors of nodes and the citation links between documents as edges in graphs. Table~\ref{Tabel_citation} summarizes the dataset statistics showing the sizes and other characteristics of the data.
\subsection{Implementation Details}
\begin{table*}
	\caption{Summary of results in terms of classification accuracy of the three datasets.}
	\centering
	\begin{tabular}{c|c|c|c|c}
		\hline
		\hline
		\multicolumn{2}{c|}{\textbf{Model}}
		& \textbf{Citeseer}
		& \textbf{Cora}
		& \textbf{Pubmed}                \\ \hline
		\multicolumn{2}{l|}{ManiReg~\cite{belkin2006manifold}}                             & 60.1\%                  & 59.5\%                  & 70.7\%                  \\
		\multicolumn{2}{l|}{SemiEmb~\cite{weston2012deep}}                             & 59.6\%                  & 59.0\%                  & 71.1\%                  \\
		\multicolumn{2}{l|}{LP~\cite{zhu2003semi}}                                  & 45.3\%                  & 68.0\%                  & 63.0\%                  \\
		\multicolumn{2}{l|}{DeepWalk~\cite{perozzi2014deepwalk}}                            & 43.2\%                  & 67.2\%                  & 65.3\%                  \\
		\multicolumn{2}{l|}{ICA~\cite{lu2003link}}                                 & 69.1\%                  & 75.1\%                  & 73.9\%                  \\
		\multicolumn{2}{l|}{Planetoid~\cite{yang2016revisiting}}                       & 64.7\%                  & 75.7\%                  & 77.2\%                  \\
		\multicolumn{2}{l|}{Chebyshev~\cite{defferrard2016convolutional}}                       & 69.8\%                  & 81.2\%                  & 74.4\%                  \\ \hline
		\multicolumn{2}{l|}{GCN~\cite{kipf2016semi}}                         & 70.3\%                  & 81.5\%                  & 79.0\%                  \\
		\multicolumn{2}{l|}{GAT~\cite{velivckovic2017graph}}                            & 72.5\%                  & 83.0\%                 & 79.0\%                 \\ \hline

		\multicolumn{2}{l|}{ GCN~\cite{kipf2016semi} \textbf{+ SSL} }
		& \textbf{72.95\% $\pm$ 0.62\%}                  & \textbf{83.80\% $\pm$ 0.73\%}                 & \textbf{81.23\% $\pm$ 0.86\% }                 \\
		\multicolumn{2}{l|}{GAT~\cite{velivckovic2017graph} \textbf{+ SSL} }
		& \textbf{72.73\% $\pm$ 0.72\% }                  & \textbf{83.70\% $\pm$ 0.61\%}                  & \textbf{79.14\% $\pm$ 0.53\% }                \\ \hline \hline
	\end{tabular}
	\label{Tabel_Results}
\end{table*}
The proposed method is implemented using the open-source deep learning library Keras~\cite{chollet2015keras}. In the SSL  phase, we first modify the structure of GCN and GAT and a link prediction layer be added in the end of each model for the link prediction task. To construct the input data of SSL, we randomly remove parts of links inside the graph and cover parts of features in the feature vectors of each node. And then train them with Adam optimizer. The L2 regularization with weight of ${\rm{5}} \times {10^{ - 4}}$ and dropout in input and hidden layers with a rate of 0.5 are also be used to overcome overfitting problem. The activation functions used for GCN and GAT are ReLU and ELU, respectively. For the GCN model, we modify the number of hidden units in the first layer from 16 to 32 to improve the representation ability. In the link prediction phase, for both datasets, we save the weights learned by the model with a patience of 5000 epochs as the final parameters weights.
In the classification phase, we first employ the parameter weights learned in the SSL phase to initialize the corresponding layer, and then fine-tune the model on the training dataset.
Finally, we report the 10-run result of the model in order to fairly evaluate the benefits of the SSL strategies.

\subsection{Effects of Self-supervised Learning}
To explore the contributions of RRL and RCF to improving the model performance, we first evaluate the performance of GCN~\cite{kipf2016semi} on the Cora dataset with three SSL strategies: independent RRL and RCF (RRL, RCF), mixed RRL and RCF (RRL \& RCF). In addition, we also evaluate the influence of the amount of removing link and covering feature on the classification task.

Table~\ref{Tabel_cora_experiment} and Figure~\ref{figure_cora_experiment} summarized the performance of GCN under SSL strategies. It's notice that we also use the 10-run result of models
as the final result. From Table~\ref{Tabel_cora_experiment} and Figure~\ref{figure_cora_experiment}, we can observe that both RRL, RCF and RRL \& RCF can improve the performance of GCN.
And compared with RRL, RCF, the model based on RRL \& RCF obtained
significantly better performance on the Cora datasets.
Apart from the evaluation criterion of accuracy, we also perform a statistical comparison of the results using paired $t$-test with a confidence interval of 0.95 to demonstrate the effectiveness of our proposed RRL and RCF. GCN with SSL is compared to GCN without SSL for statistical significance, while all $p$ values are also listed in the  corresponding Table~\ref{Tabel_cora_experiment}. It can be seen that our proposed RRL and RCF significantly outperforms the GCN without SSL strategy  with $p<0.05$, which clearly  demonstrates that our proposed SSL is effective in improving the classification performance .

We also employ t-SNE \cite{maaten2008visualizing} in our work to visualize the distribution of the node representation learned through RRL, RCF and RRL \& RCF on the link prediction task as shown in Figure~\ref{TSNE}.
It can be seen from the Figure~\ref{TSNE} that the node representation learned by our SSL is distinguished compared with input, which proves that our SSL can make the GCN models exploit and leverage the graph struct for training to further improve the classification accuracy.

\subsection{Comparisons with state-of-the-art}
We apply our proposed SSL on two representative GCN models - GCN~\cite{kipf2016semi}, GAT~\cite{velivckovic2017graph} - to demonstrate generalization as well as portability. In experiments, we use RRL\& RCF with 40\% as SSL strategies. The performances of GCN and GAT combined with the SSL strategies on the three citation network datasets are reported in Table~\ref{Tabel_Results}.
From the Table~\ref{Tabel_Results}, we can see that, under the same dataset and model, our proposed SSL strategies improve the performance of GCN by an average margin of 2.65\%, 2.30\%, 2.23\%, while the improvements are 0.23\%, 0.70\%, 0.14\% for GAT; this proves that our proposed SSL are generalized and robust to the GCN-based models. In addition, compared with other methods, GCN + SSL and GAT +SSL achieve state-of-the-art performances. This also indicates that node features can be learned by exploiting the information from the graph itself and that those node representation can be transferred to the classification task for improving model performance.

\section{Conclusions}
In this paper, to improve the performance of GCN-based models by efficiently exploiting the information in the existing data, two pretext tasks - Randomly Removing Links (RRL) and Randomly Covering Features (RCF) - are designed to introduce SSL into GCN. Both strategies exploit and leverage the graph structure for network training to further improve the performance. Extensive experiments were conducted to demonstrate the effectiveness of the proposed self-supervised training strategies. In our future work, we will investigate more information types for utilizing through learning strategies like self-supervision to further improve the network performance.
\bibliographystyle{IEEEtran}
\bibliography{newref}

\end{document}